%% file: main.tex
\documentclass[journal]{IEEEtran}
\usepackage{amsmath,amsfonts}
\usepackage{array}
\usepackage[caption=false,font=normalsize,labelfont=sf,textfont=sf]{subfig}
\usepackage{textcomp}
\usepackage{stfloats}
\usepackage{url}
\usepackage{verbatim}
\usepackage[pdftex]{graphicx}
\usepackage{balance}
\usepackage{cite}
\usepackage{multirow}
\usepackage{multicol}
\usepackage{tabularx}
\usepackage{xcolor}
\usepackage{mathtools}
\usepackage{makecell}
\usepackage[hidelinks]{hyperref}
\newif\ifreview
\reviewfalse  

\ifreview
  \newcommand{\rev}[1]{#1}
  \newcommand{\revii}[1]{{\color{blue}#1}}
\else
  \newcommand{\rev}[1]{#1}
  \newcommand{\revii}[1]{#1}
\fi

\newcommand{\orcidicon}[1]{%
  \href{https://orcid.org/#1}{%
    \raisebox{-0.15ex}{\includegraphics[height=1.6ex]{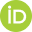}}%
  }%
}

 \usepackage{color}
 \usepackage{flushend}
\usepackage{epstopdf}
\usepackage{fancyhdr,graphicx,amsmath,amssymb}
\usepackage{algorithm}
\usepackage[noend]{algpseudocode}

\algnewcommand\algorithmicforeach{\textbf{for each}}
\algdef{S}[FOR]{Foreach}[1]{\algorithmicforeach\ #1\ \algorithmicdo}
\algdef{E}[FOR]{EndForeach}{\algorithmicend\ \algorithmicforeach}

\begin{document}
\title{Optimizing Nursing Care Taxi Dispatch \\
Leveraging Integer Linear Programming Solvers and Machine Learning\thanks{This paper is an accepted journal article on IEEE Transactions on Intelligent Transportation Systems.}}



\author{
Riku Nakao\protect\orcidicon{0009-0005-8727-5414}\IEEEauthorrefmark{1}, Akihito Hiromori\IEEEauthorrefmark{1}~\IEEEmembership{Member,~IEEE,}
Hamada Rizk\IEEEauthorrefmark{1}\IEEEauthorrefmark{2}\IEEEauthorrefmark{3}~\IEEEmembership{Senior Member,~IEEE,} \\
Hirozumi Yamaguchi\IEEEauthorrefmark{1}\IEEEauthorrefmark{2}~\IEEEmembership{Senior Member,~IEEE} \\
    \IEEEauthorblockA{\IEEEauthorrefmark{1} The University of Osaka, Japan},
\IEEEauthorblockA{\IEEEauthorrefmark{2} RIKEN R-CCS, Japan},
\IEEEauthorblockA{\IEEEauthorrefmark{3} Tanta University, Egypt} \\
    \thanks{R. N. is with Information Networking, Graduate School of Information Science and Technology, The University of Osaka, Suita 565-0871, Japan (e-mail: r-nakao@ist.osaka-u.ac.jp).}
\thanks{A. H. is with Information Networking, Graduate School of Information Science and Technology, The University of Osaka, Suita 565-0871, Japan (e-mail: hiromori@ist.osaka-u.ac.jp).}
\thanks{H. R. is with 
Information Networking, Graduate School of Information Science and Technology, The University of Osaka, Suita 565-0871, Japan, RIKEN R-CCS, Kobe 650-0047, Japan and also with the Department of Computer and Control Engineering, Tanta University, Tanta 31733, Egypt (e-mail: hamada\_rizk@ist.osaka-u.ac.jp). }
\thanks{H. Y. is with Information Networking, Graduate School of Information Science and Technology, The University of Osaka, Suita 565-0871, Japan and RIKEN R-CCS, Kobe 650-0047, Japan (e-mail: h-yamagu@ist.osaka-u.ac.jp).}
}

\markboth{IEEE Transactions on Intelligent Transportation Systems, ~Vol.~XX, No.~Y, March~2025}%
{Author1, Author2 \MakeLowercase{\textit{et al.}}: Optimizing Nursing Care Taxi Dispatch Leveraging Integer Linear Programming Solvers and Machine Learning}


\maketitle

\input{00_abstract}


\input{01_introduction}
\input{02_related_works}

\input{03_formulation}
\input{04_algorithm}
\input{05_evalation}
\input{06_conclusion}

\bibliographystyle{IEEEtran}
\bibliography{reference}

\end{document}

%% file: 00_abstract.tex
\begin{abstract} 
In this paper, we formulate a new vehicle dispatch optimization problem, called Nursing Care Taxi Dispatch, as a variant of the Vehicle Routing Problem, considering constraints related to wheelchair use, user compatibility, pick-up and drop-off times, and vehicle limitations. Previous neural-based methods for Vehicle Routing Problems have typically addressed a few simple constraints, while our new problem involves multiple complex constraints, resulting in having fewer destinations to select. This complexity makes it more difficult to obtain solutions that allow all nodes to be visited with a limited number of vehicles. To balance low violation rate, computational efficiency, and solution quality, we propose a supervised machine learning approach based on the Transformer architecture. We first obtain a set of high-quality solutions using an integer linear programming solver for given inputs and then train our \revii{learning model} through supervised learning. Additionally, we introduce the post-processing of the paths generated by the learning model, ensuring that all constraints are satisfied. \revii{We compared each instance's \revii{objective function value (operating time)}, execution time, and constraint violation rate} across different methods: our proposed method and some existing methods including integer linear programming and machine learning-based methods, using real-world facility data. Our method successfully produced balanced solutions regarding \revii{operating time}, \revii{execution time}, and constraint violation rate. \rev{Notably, we observed a decrease in the \revii{operating time} for all problem sizes and regions, while keeping constraint violations
to a minimum compared to existing methods. Especially, the decrease reached up to 8\% for problem sizes with fewer than 30 users.}
 
\end{abstract}

%% file: 01_introduction.tex
\section{Introduction}

\IEEEPARstart{T}{he} aging population in many countries \cite{wbg65} has created a critical demand for efficient and specialized transportation services (shuttle services) tailored to the unique needs of elderly individuals\cite{avta,rta}. Among these, nursing homes require reliable mobility solutions to transport residents between facilities and their homes. Such services not only involve logistical challenges but are also constrained by numerous operational requirements. These include accommodating passengers with reduced mobility, ensuring adherence to strict pick-up and drop-off schedules, and optimizing passenger grouping for both compatibility and comfort. Traditional transportation optimization methods, often designed for generic applications, fail to address the multifaceted and highly specific demands of nursing care mobility.

Providing reliable and comfortable transportation for elderly users is particularly complex due to their unique needs. Specifically, shuttle buses need to arrive at the pick-up points punctually\cite{avta}. A delayed pick-up could disrupt carefully planned schedules, cascading delays through the system and causing significant inconvenience to other users. Additionally, many elderly passengers rely on wheelchairs for mobility \cite{nie2024understanding}, requiring vehicles equipped with specialized seating and accessibility features. Moreover, users' psychological and emotional well-being must also be considered \cite{pillemer2012resident}; incompatible passengers sharing a vehicle could lead to discomfort. 
\rev{ These compatibility constraints are not limited to specialized services like nursing care transportation but can be generalized to other public transportation services such as ride hailing service. Unlike mass transit, the intimate setting of a shared vehicle often requires restricting specific passenger combinations to ensure user safety and health. For instance, separating passengers to mitigate the risk of infectious diseases is critical, particularly for vulnerable users. Furthermore, restrictions may be necessary to accommodate users with specific medical conditions or behavioral symptoms that might cause safety risks to others.
Given the frequency of conflicts in ride-sharing services such as Uber \cite{uber_report}, the implementation of systematic incompatibility constraints is indispensable. Previous research has formulated a problem that prohibits a certain pair of users from riding on the same vehicle \cite{molenbruch2017multi, schulz2024branch}. Furthermore, such incompatibility constraints have been introduced also in the healthcare domain, addressing factors such as patient-vehicle suitability and separate users with special requirements like those who need a stretcher from other passengers \cite{detti2017multi}. 
From these points of view, prohibiting co-passengers based on user compatibility represents a valid and generalizable principle for practical transportation system design.}
These constraints make it imperative to design mobility solutions that prioritize user-centric considerations while maintaining operational efficiency.

To address these challenges, this paper introduces the \textit{Nursing Care Taxi Dispatch} (NCTD) problem, a novel and complex variant of the Vehicle Routing Problem (VRP) \cite{dantzig1959truck}. The VRP is a classic combinatorial optimization problem where multiple vehicles are routed to visit a set of nodes while minimizing an objective function, typically travel cost or distance. However, unlike traditional VRP or its simpler variants, such as the Traveling Salesman Problem (TSP) \cite{applegate2006traveling} and Capacitated Vehicle Routing Problem (CVRP) \cite{clarke1964scheduling}, the NCTD problem incorporates a range of real-world constraints \revii{and defines the operating time, which is the sum of the travel time and the staying time, as the objective function.} These include user compatibility, strict pick-up and drop-off time windows, and specialized vehicle requirements such as wheelchair accessibility. 
This high-dimensional constraint space renders the NCTD problem more complex and challenging than standard VRP, making feasible solutions increasingly sparse as constraints accumulate.

Obtaining the optimal solution of VRP is NP-hard \cite{lenstra1981complexity} because VRP is a generalization of TSP \cite{dantzig1959truck}, which is also an NP-hard problem \cite{Karp1972}. This NP-hard nature of VRP  significantly increases the computational complexity of solving its advanced variants, such as the NCTD. Exhaustive search methods \cite{cordeau2006branch,ropke2009branch}, while theoretically capable of finding optimal solutions, become computationally infeasible for large-scale instances due to the exponential growth of the solution space. This limitation has driven the development of heuristic approaches, such as genetic algorithms \cite{baker2003genetic} or ant colony optimization \cite{dorigo1997ant}, as well as learning-based approaches, including Graph Neural Network (GNN) \cite{joshi2019efficient,sui2023neuralgls} or Transformer\cite{kool2018attention,kwon2020pomo}, which are designed to efficiently approximate high-quality solutions within practical time constraints.

Recent advancements in deep learning have provided promising methodologies for tackling routing problems by leveraging neural architectures to explore solution spaces effectively.  Kool et al. \cite{kool2018attention} introduced the Attention Model (AM), a method utilizing the Attention Mechanism to compute the softmax value for each node, enabling step-by-step generation of solutions. This approach incorporates reinforcement learning (RL) to maximize the probability of selecting routes with a minimal total travel distance, thereby optimizing performance. Building on this foundation, Kwon et al. \cite{kwon2020pomo} facilitate the identification of the optimal solution for VRP instances by exploring feasible solutions from multiple initial nodes.
Similarly, Bi et al. \cite{bi2022learning} employed knowledge distillation techniques to enhance \revii{the generalization of the learning model} for VRP instances with diverse demand distributions.
While these methods demonstrate promising results for conventional VRP variants, such as the TSP and the CVRP, they are not well-suited for addressing complex, constraint-heavy problems like NCTD. The intricate dependencies and real-world constraints of NCTD, such as user compatibility, wheelchair accessibility, and strict pick-up and drop-off windows, significantly reduce the feasible solution space. Consequently, existing approaches struggle to produce high-quality solutions under these conditions, especially when vehicle availability is limited.

To overcome the challenges posed by the complex constraints and operational requirements of the NCTD problem, we propose a novel hybrid solution that integrates linear programming, supervised learning, and heuristic refinement. Our approach begins by formulating the NCTD problem as a mixed-integer linear programming (MILP) model (Section \ref{sec:Formulation}), explicitly incorporating real-world constraints such as vehicle capacity, pick-up and drop-off schedules, and user-specific requirements, including wheelchair accessibility and passenger compatibility. This formulation ensures a comprehensive representation of the problem, enabling the generation of high-quality feasible solutions that serve as a foundation for further processing.
Then, we develop a supervised learning framework grounded in the AM \cite{kool2018attention}. There are mainly three key innovations in this framework. 
Firstly, the framework learns to mimic the feasible routing patterns, or ground truth labels of the next destinations, generated by the MILP model, preventing it from producing low-quality solutions (Section \ref{sec:training}).
Secondly, the framework leverages contextual embeddings that encode spatial, temporal, and personal information such as accessibility and compatibility, further enhancing its ability to adapt to diverse NCTD instances (Section \ref{sec:arch}).
Lastly, especially in the inference phase, we expand state-aware masking mechanisms suggested in AM\cite{kool2018attention} to the NCTD, which dynamically restricts the \revii{learning model's} decision space to valid actions based on the current state of the vehicle and route. This capability allows \revii{the learning model} to generate almost feasible solutions efficiently while adhering to real-time constraints (Section \ref{sec:masks}).
After the learning process, we introduce a heuristic enhancement step inspired by the Insertion Algorithm \cite{rosenkrantz1977analysis} to further refine the solutions produced by the learning framework (Section \ref{sec:ArrangingTours}).
This process is indispensable since the constraint of limited vehicle availability is not fully addressed during the learning process. This heuristic reorganizes routes when the learning-based solution does not satisfy all constraints and ensures strict compliance with all constraints, which plays a significant role in real-world facilities where the number of vehicles is limited.
%
By combining MILP-based optimization, Transformer-based learning, and heuristic refinement into a unified system, our approach provides a robust and scalable framework for solving complex, constraint-heavy routing problems. Also, it establishes a new benchmark for tackling real-world transportation challenges, particularly in nursing care mobility.

To validate the effectiveness of our proposed approach, we conducted extensive evaluations using NCTD instances with varying complexities, including 20, 30, and 50 nodes. These instances were generated from real-world pickup data provided by two nursing care facilities and solved using an ILP solver to produce high-quality training data for the supervised learning framework. \revii{After the learning model was trained, we performed inference and obtained the final solutions. This method was evaluated against the ILP solver (IBM CPLEX and HiGHS\cite{huangfu2018parallelizing}) and state-of-the-art deep learning methods based on metrics such as \revii{operating time, execution time,} and constraint violation rates.}
\rev{Our approach demonstrated a consistent reduction in the \revii{operating time} across instances of all sizes and regions, achieving reductions of up to 8\% for instances with 30 or fewer nodes. The method achieved zero constraint violations in all but one instance. Furthermore, \revii{the execution time of our method was comparable to existing methods in most cases}, although some instances required longer solution times, ensuring practical feasibility for real-world deployment. These results highlight the effectiveness of our hybrid solution in addressing the unique challenges of the NCTD problem, paving the way for its application in efficient and reliable nursing care transportation systems.}

%% file: 02_related_works.tex
\section{Related Works}
In recent years, various extensions of the VRP have been proposed, incorporating additional constraints to address real-world challenges. Some of these extensions focus on mobility assistance for older adults. For instance, the Dial-a-Ride Problem (DARP) \cite{cordeau2006branch, ropke2007models} introduces a two-point constraint for each user (a start and a destination) along with time-window constraints for arrival. Liu et al. \cite{Liu2018} formulate further constraints reflecting real-world service scenarios, such as allowable time ranges and the duration of services at users' homes. Additionally, Parragh \cite{parragh2011introducing} considers heterogeneous vehicle configurations, such as wheelchair-accessible seating and stretcher accommodations, to address the specific needs of users in mobility assistance scenarios.

\subsection{Exact Methods}
Exact methods are widely used for solving combinatorial optimization problems like the VRP as they provide guaranteed optimal solutions. Among these, Branch-and-Bound (B\&B)\cite{Lee_2004} is a foundational technique that systematically reduces the solution space. In B\&B, the original minimization (or maximization) problem $P$ is divided into smaller subproblems $P_i(i=1,2,\dots)$ by fixing certain variables. If the lower (or upper) bound of a subproblem $P_i$ exceeds the best-known solution of $P$, further exploration of $P_i$ is unnecessary. This pruning mechanism significantly improves computational efficiency by focusing only on promising regions of the solution space.

Over the years, enhanced versions of B\&B have been developed to address the growing complexity of optimization problems. One such approach is the Branch-and-Cut (B\&C)\cite{Lee_2004} method, which combines B\&B with cutting-plane methods to refine the feasible region of the linear relaxation of the problem. This refinement introduces additional constraints, or "cuts," that eliminate infeasible solutions without removing any optimal ones. Studies such as \cite{cordeau2006branch, ropke2007models} demonstrate the effectiveness of B\&C in solving the DARP, where users are transported between specific locations under time and capacity constraints. For example, Ropke and Cordeau \cite{ropke2009branch} applied B\&C with tailored constraints to efficiently solve DARP instances involving up to 200 nodes.

Another advancement in exact methods is the Branch-and-Price (B\&P)\cite{Lee_2004} technique, which integrates B\&B with column generation, a method particularly useful for problems with a large number of variables. In this approach, only a subset of variables is considered initially, and additional variables are iteratively introduced when they improve the solution. This approach is especially effective for sparse problems, where many variables have zero coefficients in the optimal solution. Applications of B\&P in \cite{ropke2009branch,garaix2011optimization} highlight its ability to solve large-scale vehicle routing problems efficiently.
Reinhardt et al. \cite{reinhardt2016edge} extend the B\&P framework by incorporating cutting-plane techniques, resulting in the Branch-and-Price-and-Cut (BP\&C) method. This hybrid approach has proven particularly effective for Vehicle Routing Problems with Time Windows (VRPTW), achieving optimal solutions for instances involving up to 40 users within a reasonable computational time. By combining the strengths of branching, pricing, and cutting, BP\&C represents a state-of-the-art exact method for handling complex constraints in routing problems.

While exact methods such as B\&B, B\&C, and B\&P guarantee optimality, their computational demands often grow exponentially with problem size. For instance, the evaluation of numerous subproblems or the introduction of extensive numbers of variables and cuts makes these methods impractical for large-scale, real-world scenarios where time constraints are critical. This computational overhead limits the scalability of exact methods, underscoring the need for heuristic or approximate approaches that balance solution quality and computational efficiency.

\textit{On the contrary, the proposed approach addresses the computational challenges of exact methods by combining optimization-based modeling with learning and refinement strategies. Instead of directly solving each instance, we use an ILP solver only to generate feasible and high-quality solutions for training a learning framework that generalizes constraints and patterns. Unlike traditional exact approaches, our method employs dynamic state-aware adjustments to ensure the feasibility of solutions during generation, while a refinement stage enhances robustness and resolves inconsistencies. This hybrid method balances the precision of optimization with the efficiency and scalability needed for real-world, constraint-heavy problems like NCTD.}

\revii{\subsection{Evolutionary Algorithms}
Traditional optimization methods include evolutionary algorithms, such as the Genetic Algorithms (GA). In GA, a fixed number of individuals representing potential solutions are maintained, and new individuals are generated through genetic operators such as crossover or mutation. The algorithm then searches for solutions by retaining high-quality individuals for the next generation via selection based on objective function values. As described in Chapter \ref{sec:Formulation}, the NCTD addressed in this paper has multiple objective functions; therefore, we can apply multi-objective optimization methods to NCTD. Well-known representative methods include the Non-dominated Sorting Genetic Algorithm II (NSGA-II) \cite{deb2002fast} and the Multi-objective Evolutionary Algorithm based on Decomposition (MOEA/D) \cite{zhang2007moea}. In multi-objective optimization, trade-offs generally exist between multiple objective functions; consequently, the aim is to find a set of solutions (the Pareto front) where improving one objective worsens another. In NSGA-II, solutions are ranked using non-dominated sorting based on their closeness to the Pareto front, while crowding distance is used to preserve the diversity of the solutions. Furthermore, in MOEA/D, multi-objective optimization problems are decomposed into multiple single-objective optimization problems, and the sub-problems defined for different weight vectors are solved in parallel to efficiently approximate the Pareto front. Evolutionary algorithms can be also applied to routing problems; for example, in \cite{li2025optimization}, a genetic algorithm called ``nondominated sorting genetic algorithm with mass center (NSGA-MC)'' is proposed to address a multi-objective routing optimization of the robotaxi dispatch.
}

\subsection{Machine Learning-based Methods} \label{sec:MLbased}
Machine learning has emerged as a powerful approach to solving routing problems such as the TSP and VRP, particularly in addressing the limitations of exact methods for large-scale or complex scenarios. These approaches can be broadly categorized into supervised learning and RL, each offering unique advantages and challenges.

Supervised learning methods rely on high-quality ground-truth datasets to train \revii{learning models} that approximate optimal or near-optimal solutions. A notable example is the Pointer Network \cite{vinyals2015pointer}, which maps input coordinates to output permutations of nodes to solve TSP. The flexibility of handling variable input sizes makes the Pointer Network adaptable to problems of different scales. However, it requires the input and output sequence lengths to be equal, which limits its applicability to VRP, where revisiting nodes is possible. Milan et al. \cite{milan2017data} address this limitation by employing recurrent neural networks (RNNs), demonstrating superior performance for NP-hard problems like TSP.
Further advancements in supervised learning leverage GNNs and Graph Convolutional Networks (GCNs). These approaches effectively capture spatial and relational features of routing problems. For instance, Joshi et al. \cite{joshi2019efficient} utilize GCNs to predict the probability of edge inclusion in optimal solutions, achieving an error of only 0.01\% for 50-node TSP instances. Additionally, methods combining GCNs with Guided Local Search (GLS) have shown promise. Sui et al. \cite{sui2023neuralgls} estimate edge "regret" values, representing their contribution to the solution's cost. By identifying and removing edges with high regret, GLS avoids suboptimal solutions and local optima, demonstrating robust performance across various routing scenarios.
\rev{ In addition, some papers tackle large-scale problems. 
 Li et al. \cite{li2021learning} segments large problems into smaller sub-problems and optimizes how to segment them through learning, and Luo et al. \cite{luo2023neural} propose an embedding approach that intentionally limits the capture of detailed graph structure using the Attention mechanism to enhance generalization, and applies supervised learning to improve learning efficiency.}

RL methods, on the other hand, optimize routing policies without requiring explicit ground-truth solutions. These methods are particularly advantageous for problems with dynamic or incomplete information. The REINFORCE algorithm \cite{williams1992simple} is widely used to update \revii{learning model} parameters by maximizing cumulative rewards while minimizing the variance of gradient estimates. Kool et al. \cite{kool2018attention} employ an Attention Mechanism to probabilistically select the next destination based on vehicle and node attributes, such as position and capacity. By iteratively optimizing attention parameters, this approach minimizes total travel distance and achieves high-quality solutions for various VRP variants.
Expanding on these ideas, Kwon et al. \cite{kwon2020pomo} introduce a parallelized RL method that generates multiple solutions simultaneously by treating different nodes as starting points. This parallel approach improves the likelihood of finding near-optimal solutions efficiently. 
\rev{ Based on AM, several methods have been proposed to deal with problems with complex constraints. Gao et al. \cite{gao2023towards} includes not only the traditional policy that determines the next destination from the entire nodes, but also a policy that focuses only on nearby nodes, which results in capturing a policy that does not depend on the distribution of nodes and the problem size. Additionally, other methods reflect constraint violations in the learning process by including them as input features \cite{ma2023learning} or objective functions \cite{bi2024learning}.}
Another common RL strategy involves Sample Rollout and Greedy Rollout. Sample Rollout explores diverse solutions by probabilistically selecting the next node, while Greedy Rollout deterministically selects the highest probability node to exploit known policies. However, as noted in \cite{kool2018attention, peng2020deep, kwon2020pomo}, Greedy Rollout often reduces solution quality by limiting exploration, particularly during the early stages of training when policies are less refined.

\revii{Studies for recent years have also shown that learning-based methods can effectively solve classical routing problems with relatively simple constraints, such as TSP and CVRP. Representative examples include frameworks that iteratively refine solutions through divide-and-conquer strategies \cite{zheng2024udc}, unsupervised models that directly learn feasible high-quality tours for TSP\cite{min2023unsupervised}, and neural approaches that guide local search operators such as 2-opt\cite{guan2025synergetic}. Moreover, some problem-specific architectures have been proposed for problems like Electric VRP, which is a problem with constraints specific to electric vehicles.\cite{wang2025attention}}

Despite significant progress, machine learning approaches face notable challenges. Supervised learning methods depend heavily on the availability of high-quality ground-truth solutions.
RL-based methods, while eliminating the need for explicit ground truth, often struggle with balancing exploration and exploitation, leading to slower convergence. Furthermore, the performance of both approaches is highly sensitive to neural architecture design and problem-specific adaptations, such as handling multiple constraints and \revii{dynamic inputs, and problem-specific structures\cite{min2023unsupervised,zheng2024udc,wang2025attention} or search operators\cite{guan2025synergetic}}, making their application to complex VRP variants, like NCTD, non-trivial.

\textit{On the contrary, the proposed approach combines the strengths of supervised learning and heuristic refinement while leveraging optimization-based formulations. By generating high-quality training data from an ILP solver, the \revii{learning model is trained} to capture intricate routing patterns and constraint satisfaction without the computational overhead of directly solving these instances repeatedly and the long learning process that is noticeable particularly in RL. Furthermore, the inclusion of state-aware mechanisms ensures dynamic feasibility during solution generation, addressing the limitations of both supervised and RL methods in handling complex, constraint-rich problems like NCTD.}

%% file: 03_formulation.tex
\section{Formulation} \label{sec:Formulation}
\begin{table}[t]
    \centering
    \caption{Constants of Nursing Care Taxi Dispatch}
    \begin{tabular}{ll}
    \hline
    \textbf{Notation}  & \textbf{Description}  \\
    \hline
    $N_U \in \mathbb{N}$     & The number of total users \\
    $U=\{1,..,N_U\}$ & the set of users \\
    $U_0=\{0\} \cup U$ & the set of users and the facility \\
    $N_V \in \mathbb{N}$ & the available number of vehicles \\
    $V=\{1,..,N_V\}$ & the set of vehicles \\
    $A_i$ & the attributes associated to user $i$ \\
    $loc_{i}=(\text{lat}_i,\text{lng}_i)$ & the location(latitude and longitude) of user $i$'s house \\
    $loc_{0}=(\text{lat}_0,\text{lng}_0)$ & the location(latitude and longitude) of the facility \\
    $T_{f} \in \mathbb{Z}$ &  the end time of the staying \\
    $T^S_i \in [0,T_f]$ & user $i$'s expected departure time from his/her house  \\
    $T^G_i \in [0,T_f]$ & user $i$'s expected arrival time to the facility \\
    $T_{i}^{\text{task}} \in [0,T_f]$ & the time needed to perform tasks at user's house \\
    \rev{$inc_{i} \in \{0,1\}^{N_U}$} & \rev{vector representing user $i$'s incompatibility} \\ 
    \rev{ $N_I$} & \rev{ the number of incompatible pairs} \\
    $wheel_{i} \in \{0,1\}$ & user $i$'s wheelchair usage in the vehicle \\
    $c_{i,j} \in \mathbb{R}_{>0}$ & the time required to pass between $loc_i$ and $loc_j$ \\
    $cap_{k}^{W}, cap_{k}^{N} \in \mathbb{Z}$ & the number of wheelchair seats / normal seats \\
    $M$ & a large constant used in Time Constraint \\
    $\epsilon$ & an infinitesimal quantity \\
    \hline
    \end{tabular}
    \label{tab:notation_table}
\end{table}
\begin{table}[t]
    \centering
    \caption{parameters related to the time constraint}
    \begin{tabular}{ll}
        \hline
        \textbf{Parameter} & \textbf{Description} \\ \hline
        $\alpha$ & an allowable window size to the user's pickup time \\
        $\beta$ & a buffer time before the expected arrival time \\ 
        $\gamma$ & the maximum waiting time at each house\\
        \hline
    \end{tabular}
    \label{tab:timeparam}
\end{table}
\begin{table}[t]
\centering
\caption{Variables in Nursing Care Taxi Dispatch}
\label{tab:notation_table_val}
\begin{tabular}{ll}
\hline
\textbf{Notation} & \textbf{Description} \\
\hline
\multicolumn{2}{l}{\textit{Main decision variables}} \\[2pt]
$ x^k_{i,j} $ & 1 if vehicle $k$ travels from $i$ to $j$ \\
$ t_i $ & Pickup time at user $i$'s house \\
$ t^k_0 $ & Time when vehicle $k$ returns to the facility \\
$ z $ & Auxiliary variable representing the maximum time \\
\rev{$ s^+_{j,k}, s^-_{j,k} $} & \rev{Auxiliary variables used for linearization of staying time} \\[4pt]
\hline
\multicolumn{2}{l}{\textit{Derived variables}} \\[2pt]
\rev{$ \tilde{x}_{i,j} $} & \rev{1 if any vehicle traverses from $i$ to $j$,} \\
& \rev{defined as $\tilde{x}_{i,j} = \sum_{k=1}^{N_V} x^k_{i,j}$} \\
\rev{$ \hat{x}_{j,k} $} & \rev{1 if user $j$ is assigned to vehicle $k$,} \\
& \rev{defined as $\hat{x}_{j,k} = \sum_{i=0}^{N_U} x^k_{i,j}$} \\
\hline
\end{tabular}
\end{table}

We formulate the NCTD problem as an extension of VRP by incorporating additional constraints specific to transportation services for older adults.
In the traditional VRP, multiple vehicles start from a central depot, visit designated nodes, and return to the depot. Each node is visited exactly once by one vehicle, and the objective is to minimize the total travel length while satisfying all constraints. In the context of NCTD, we redefine the depot as a "facility", nodes as "users", and visitors as "vehicles" to align with the caregiving scenario.

In this paper, we define four constraints unique to NCTD in addition to those originally included in VRP to simulate practical caregiving scenarios. First, \textit{Vehicle Number Constraint} accounts for the limited availability of vehicles in real-world facilities, restricting the total number of vehicles that can be dispatched. Second, \textit{Capacity Constraint} ensures vehicles comply with specific seating capacities, including normal and wheelchair-accessible seats, to accommodate users with different physical needs. Wheelchair seats are reserved for users requiring additional support, ensuring their comfort during transportation. Third, \textit{Time Constraint} enforces that vehicles pick up users within their requested time windows and return to the facility by a designated time. These constraints reflect real-world considerations, such as minimizing delays to avoid user dissatisfaction and meeting facility schedules for organized events. Lastly, \textit{Incompatibility Constraint} prohibits  facility-defined incompatible pairs from sharing the same vehicle. This constraint is crucial for ensuring users' safety, such as by mitigating infection risks and facilitating driver supervision of passengers, particularly those requiring significant assistance. \rev{ Under these constraints, the overall objective is to minimize the sum of the average and the maximum operating time across all vehicles, defined as the sum of travel time and staying time.}


\subsection{Input Values and Variables}
We also introduce several NCTD-specific constants.
All the relevant constants are summarized in Table \ref{tab:notation_table}. First, we set the available number of vehicles $N_V$ to apply the vehicle number constraint. Moreover, we have each user's attribute $A_{i} (i=1,\dots,N_U)$ which is represented as a $N_A$-tuple:
\begin{equation}
    A_{i} = (A_{i,1},\dots,A_{i,N_{A}})
\end{equation}
where $A_{i,j}(j=1,\dots,N_{A})$ is $j$-th attribute of user $i$.
The NCTD problem has following six attributes to user $i$: the location of the house $loc_{i}$, the expected departure time $T_i^{S}$, the expected arrival time $T_i^{G}$, the task time at users' houses $T_i^{\text{task}}$, the incompatibility vector $inc_i$, and wheelchair usage $wheel_{i}$. Thus, $A_i$ can be represented as $A_i = (loc_i, wheel_{i}, T_i^S, T_i^G, T_i^{\text{task}}, inc_i) (N_A=6)$.
$wheel_{i}$ is used to judge whether user $i$ needs to be seated in a wheelchair seat in the capacity constraint. 
Constants related to the time constraint include $T_i^S, T_i^G, T_i^{\text{task}}$, whose units are all in minutes (min). As for time constants, since the operation hours vary from facility to facility, we set the beginning time of the operation as 0 min and the end time as $T_f$ min. We defined $T_{i}^{\text{task}}$ because it takes time to do various tasks occurring at users' houses, such as parking by the house or helping them get into the car. 
The constant $inc_{i}$ represents user $i$'s incompatibility to others and denoted as $N_U$-dimensional binary vector: $inc_i=(inc_{i,1},\dots,inc_{i,N_{U}})$, whose $j$-th component is 1 if user $i$ and user $j$ cannot be assigned to the same vehicle. 
The travel time between two users' houses $c_{i,j}$ is calculated in advance. Although this value varies due to various factors such as traffic or weather conditions, we assume a constant speed for simplicity in the problem. Specifically in this paper, we calculate $c_{i,j}$ with the vehicle speed of 40km/h as a representative speed when passing roads of various widths.
We also have vehicle attributes. Every vehicle has a different number of wheelchair seats $cap_k^W$ and normal seats $cap_k^N$ $(k=1,\dots,N_V)$.

NCTD incorporates three types of decision variables, as shown in Table \ref{tab:notation_table_val}.
The variable $x_{i,j}^{k}$ represents which two users vehicle $k$ passes between, and its adjacency matrix is denoted by $X_k=[x_{i,j}^{k}]$. The variable $t_i$ represents the pickup time at user $i$'s house and $t_0^k$ is the time when vehicle $k$ gets back to the facility. Both of these are needed to check the satisfaction of the time constraint. \rev{ The auxiliary variables $z, s_{j,k}^+, s_{j,k}^-$ are used to formulate linear expressions. We denote adjacency matrices of $s_{j,k}^+,s_{j,k}^-$ as $S^+,S^-$. Moreover, we introduce two derived variables to simplify the expression: $\tilde{x}_{i,j}$ and $\hat{x}_{i,j}$.}

\subsection{Formulation of NCTD Constraints}
  In this section, we formulate the NCTD constraints as a mixed-integer linear programming model. This formulation serves two purposes: (1) to generate high-quality labels for training the supervised learning model described in Section \ref{sec:Method}, and (2) to provide a benchmark for evaluating the performance of machine learning models in Section \ref{sec:Evalation}.

\subsubsection{Vehicle Number Constraint}
\rev{ This constraint ensures that the number of vehicles used in the solution does not exceed a pre-determined maximum number of vehicles $N_V$. For this constraint, in ILP, since we declare variables based on the maximum number of vehicles, there is no risk of the number of vehicles being exceeded before solving and we do not need to explicitly impose constraint expressions containing variables. On the other hand, in ML-based Architecture introduced in Section \ref{sec:Method}, the number of vehicles cannot be determined until the last user is visited, due to the step-by-step nature of generating routes. Therefore,  unlike ILP, the constraint is not satisfied at the problem design and must be modified while generating solution. In this paper, post-processing, \revii{introduced} in Section \ref{sec:ArrangingTours}, is used to adjust the number of vehicles if the number of vehicles is exceeded after executing \revii{the learning model}.}

\subsubsection{Capacity Constraint} 
\label{sec:capConst}
The capacity constraint ensures that each vehicle adheres to its seating limitations, including both normal seats and wheelchair-accessible seats. 
\rev{ We introduce a constant $wheel_j$, which indicates whether user $j$ requires a wheelchair-accessible seat ($wheel_j=1$) or not ($wheel_j=0$). The variable $x_{i,j}^k$ is set to 1 if vehicle $k$ travels from user $i$ to user $j$, and 0 otherwise. 
Thus, the product $wheel_j  x_{i,j}^k$ equals 1 if user $j$ uses a wheelchair and is picked up by vehicle $k$, while $(1-wheel_j)x_{i,j}^k$ is 1 if user $j$ does not use a wheelchair and is picked up by vehicle $k$. Adding these terms over all users on vehicle $k$'s entire route gives the total number of wheelchair and non-wheelchair passengers, respectively. The capacity constraint requires that neither the total number of wheelchair passengers nor the total number of non-wheelchair passengers assigned to a vehicle exceed its corresponding capacity. Therefore, the constraint equation can be expressed as follows:}

\begin{align}
    \forall k \in V \left[ \sum\nolimits_{i=0}^{N_U} \sum\nolimits_{j=1}^{N_U} wheel_j x_{i,j}^k \leq cap_k^W \right] \label{eq:wheel} \\
    \forall k \in V \left[ \sum\nolimits_{i=0}^{N_U} \sum\nolimits_{j=1}^{N_U} (1 - wheel_j) x_{i,j}^k \leq cap_k^N \right] \label{eq:normal} 
\end{align}
Equation (\ref{eq:wheel}) ensures that the total number of wheelchair users assigned to vehicle $k$ does not exceed its wheelchair seat capacity $cap_k^W$. Similarly, Equation (\ref{eq:normal}) ensures that the total number of non-wheelchair users does not exceed the normal seat capacity $cap_k^N$.
These constraints are essential for maintaining the feasibility of vehicle assignments while accommodating the specific needs of users. Each of the two types of seats has a different maximum capacity as shown in Table \ref{tab:notation_table}.

\subsubsection{Time Constraint} \label{sec:timeConst}
The Time constraint consists of three smaller constraints: departure time constraint, arrival time constraint, and visiting restriction constraint.

Firstly, the departure time constraint guarantees that vehicles visit users at their desired times and ensures that users do not experience excessive waiting times.
\rev{ The constant $T_j^\text{task}$ represents the staying time required to pick up user $j$, and the constant $c_{i,j}$ denotes the travel time between nodes $i$ and $j$. When a vehicle travels from user $i$ to user $j$, i.e., when $\tilde{x}_{i,j} = 1$, the arrival at user $j$ must occur after the pickup task at $i$ is completed and the travel is finished. This relationship is expressed as:
\begin{equation}
t_i + T_i^{\text{task}} + c_{i,j} \leq t_j \label{eq:tmp_depart}    
\end{equation}
On the other hand, if no vehicle travels from user $i$ to user $j$, i.e., when $\tilde{x}_{i,j}=0$, the constraint should not restrict the value of $t_j$. Using the big-M method, which is a common technique in TSP to eliminate subtours \cite{miller1960integer}, we incorporate both cases by adding  $-M(1- \tilde{x}_{i,j})$ on the left-hand side of (\ref{eq:tmp_depart}).
Thus, the departure time constraint can be formulated as follows:
\begin{equation}
    (\forall i \!\in\! U_0)(\forall j \!\in\! U)\bigg[t_i \!+\! T_i^{\text{task}} \!+\! c_{i,j}  \!-\! M(1 \! - \! \tilde{x}_{i,j}) \!\leq\! t_j \bigg] \label{eq:timedep1}
\end{equation}
Furthermore, users' pickup time should fall within each user's preferred time window, based on the desired departure time $T_j^S$ and the allowable deviation $\alpha$. This is ensured by the following constraint:}
\begin{equation}
    T_j^S - \alpha \leq t_j \leq T_j^S + \alpha \label{eq:timedep2}
\end{equation}
where $\alpha$ is set to 10 (minutes). 
We note that (\ref{eq:timedep1}) eliminates subtours, which is also required in the formulation of TSP \cite{miller1960integer}. Subtours are paths that do not include the facility, and a trajectory including subtours is infeasible as a solution to the NCTD. A constant $M$ in (\ref{eq:timedep1}) is a sufficiently large constant (we set it to 300 minutes as a sufficiently large number). It follows that for solutions satisfying (\ref{eq:timedep1}), the pickup time $t_i$ increases monotonically except for the edge from a user to the facility when we follow the directed edges of the graph. If a subtour exists, the monotonicity of $t_i$ no longer holds between two points, except for the facility. Therefore, (\ref{eq:timedep1}) is necessary to eliminate subtours.

Secondly, arrival time constraint helps users arrive at the facility by a specified time, for example, when an event is held. 
\rev{ We use the variable $t_k^0$ to represent vehicle $k$'s arrival time at the facility.
The inequality for determining the lower bound of the facility arrival time leverages the big-M method, similar to the departure time constraint. We divide into cases according to $x_{i,0}^k$, which determines whether there is a vehicle traveling to the facility.
Since the vehicle must depart from user $i$ after completing the service $T_i^\text{task}$ and travel to the facility $c_{i,0}$, the equation is as follows:
\begin{equation}
    (\forall i \in U)(\forall k \in V) \; \left[ t_i \!+\! T_i^{\text{task}} \!+\! c_{i,0} \!-\! M(1 \!-\! x_{i,0}^k) \!\leq\! t_k^0 \right] \label{eq:timearr1}
\end{equation}
For the inequality for determining the upper bound of the arrival time at the facility, we use a derived variable $\hat{x}_{j,k}$. To help them avoid being late, we introduce a buffer time $\beta$ (Table \ref{tab:timeparam}), which allows users to arrive at the facility a little early. If user $j$ is assigned to vehicle $k$, the vehicle should arrive at the facility by $T_j^{G}-\beta$.  Therefore, the equation can be formulated as follows:
\begin{align}
(\forall j \in U)(\forall k \in V) \; & \left[ t_k^0 \leq T_j^G - \beta + M(1 - \hat{x}_{j,k}) \right]  \label{eq:timearr2}
\end{align}
where $\beta$ is 5 (minutes).
}

Lastly, visiting restriction constraint prevents users from being selected as the next destination if their expected departure times are far apart from the currently visited user. 
\rev{ The difference in scheduled pickup time between user $i$ and user $j$ is given by $|T_j^S - T_i^S|$. We must take at least $T_i^{\text{task}}+c_{i,j}$ to travel from user $i$ and user $j$. To this, to avoid burdening passengers with a long waiting time, we add parameter $\gamma$ (Table \ref{tab:timeparam}), which represents the maximum allowing waiting time at each house. 
The constraint requires that the pickup time difference between two points does not exceed the sum of the travel time, task time, and $\gamma$. Therefore, we can describe the inequality as follows:
\begin{equation}
    \tilde{x}_{i,j}=0 \quad \text{whenever }  |T_j^S - T_i^S| > T_i^{\text{task}} + c_{i,j} + \gamma .\label{eq:timerest}
\end{equation}
}
where $\gamma=30$ (minutes). 
\subsubsection{Incompatibility Constraint} \label{sec:Prefconst}
\rev{
The incompatibility constraint ensures that specific pairs of users, designated by the facility as incompatible for safety or caregiving reasons, are not assigned to the same vehicle. This incompatibility is indicated by a binary constant $inc_{i,j}$, where $inc_{i,j}=1$ indicates that users $i$ and $j$ must not ride together. We use the variable $\hat{x}_{j,k}$ to represent the assignment of users to vehicles. If and only if $\hat{x}_{i,k}+\hat{x}_{j,k}=2$, users $i$ and $j$ are assigned to the same vehicle. Since this sum must not reach 2 for any incompatibility pair of users, the incompatibility constraint is formulated as follows:
\begin{equation}
     (\forall k \in V)[\hat{x}_{i,k} + \hat{x}_{j,k} \leq 1] \quad \text{whenever } inc_{i,j} = 1.\label{eq:pref_milp}
\end{equation}
}
\subsubsection{Original Constraints in VRP}
\rev{
In a traditional Vehicle Routing Problem, the indegree and outdegree for every user must be 1, guaranteeing that every user is visited exactly once by a single vehicle. 
The inflow to user $j$ can be described as the sum of $x_{i,j}^{k}$ over all $i$ and $k$. Similarly, the outflow from user $i$ can be described as the sum of $x_{i,j}^{k}$ for all $j$ and $k$. Both of these values must be 1 if a valid route is to be obtained:
\begin{align}
    \forall j \in U \left[ \sum\nolimits_{i=1}^{N_U}\tilde{x}_{i,j} = 1 \right] \label{eq:in} \\
    \forall i \in U \left[ \sum\nolimits_{j=1}^{N_U}\tilde{x}_{i,j} = 1 \right] \label{eq:out} 
\end{align}

Moreover, every vehicle must enter and exit the facility exactly once, ensuring that all vehicles are utilized. For each vehicle $k$, the number of departures from the facility is the sum of $x_{0,j}^{k}$ over all $j$, and the number of arrivals at the facility is the sum of $x_{i,0}^{k}$ for all $i$. Both of these sums must equal 1. Thus, the following expressions must hold:
\begin{align}
    \forall k \in V \left[ \sum\nolimits_{i=1}^{N_U} x_{i,0}^{k} = 1 \right] \label{eq:fac_in}\\
    \forall k \in V \left[ \sum\nolimits_{j=1}^{N_U} x_{0,j}^{k} = 1 \right] \label{eq:fac_out}
\end{align}
}

\rev{
Lastly, vehicles must neither emerge unexpectedly from any node other than the facility nor vanish immediately after entering a node. In other words, once a vehicle visits a node, it must also depart from it. This requirement is ensured by the following equation:
\begin{equation}
    (\forall j\in U)(\forall k \in V) \left[ \sum\nolimits_{i=0}^{N_U}x_{i,j}^{k}=\sum\nolimits_{i=0}^{N_U}x_{j,i}^k\right]
\end{equation}
}

\rev{
\subsection{The problem's Complexity}
\begin{table}[t]
    \centering
    \caption{The number of each constraints}
    \begin{tabular}{l|l}
        \hline
        \textbf{The type of the constraint} & \textbf{The number of expressions} \\ \hline
        Capacity Constraint & $2N_V$ \\
        Time Constraint & $N_U^2 + 2N_UN_V$\\
        Incompatibility Constraint & $N_VN_I$ \\ \hline
    \end{tabular}
    \label{tab:number_of_constraints}
\end{table}
We show each constraint's complexity in Table  \ref{tab:number_of_constraints}. 
Of the three types of constraints, the Time Constraint has the most significant impact on the complexity of the problem, with its principal term being proportional to the square of the number of users ($N_U^2$). Furthermore, the Incompatibility Constraint depends on the number of incompatible pairs ($N_I$), making the problem more computationally challenging as there are more incompatible pairs. Thus, since the problem includes constraints that scale with the number of users, the total number of constraint expressions tends to grow readily.
}
\subsection{Objective Function}
We define the operating time as the sum of travel time and staying time. At this time, NCTD aims to minimize the following two values: (1) the mean of the operating time for all vehicles (MeanOT for short) and (2) the maximum operating time for all vehicles(MaxOT for short).
Minimizing the MeanOT reduces the overall burden on both drivers and users. Moreover, minimizing MaxOT prevents a specific car from driving for a long time. 
\rev{
We define the objective function, denoted as $f(\cdot)$ for simplicity, based on the arguments  $X_1,...,X_{N_V},S^+,S^-$. This function is designed to minimize the two aforementioned metrics: MeanOT ($f_{\text{mean}}(\cdot)$) and MaxOT ($f_{\text{max}}(\cdot)$).
Both $f_{\text{mean}}(\cdot)$ and $f_{\text{max}}(\cdot)$ are derived from the operating time (OT) of each individual vehicle $k$. The OT for vehicle $k$ ($OT_k$) is calculated as a weighted sum of its total travel time $f_{\text{travel}}^k(X_k)$,representing the sum of the edge costs $c_{i,j}$ for all edges $(i,j)$ traversed by vehicle $k$, and its total staying time $f_{\text{stay}}^k(X_k, S^+, S^-)$, representing the cumulative time vehicle $k$ spends at the visited users' locations. The staying time at user $j$'s home can be expressed as $t_j - (t_i + T_{i}^{\text{task}} + c_{i,j}) + T_j^{\text{task}}$ (or $T_j^\text{task}$ if the previous visit point is the facility), considering the staying time at the previous point and the travel time between two points.
We assign weight coefficients $w_t$ and $w_s$ to the travel and staying times, respectively, to control their relative importance.
Thus, vehicle $k$'s operating time can be represented as below:
\begin{equation}
    OT_k = w_tf_{\text{travel}}^k(X_k) + w_sf_{\text{stay}}^k(X_k, S^+, S^-)
\end{equation}
In this study, we set $w_t=0.9$ and $w_s=0.1$ to emphasize travel efficiency. 
Similarly, weights are also assigned to MeanOT and MaxOT, denoted as $w_{\text{mean}}, w_{\text{max}}$ respectively. Since both metrics are equally important for our problem, we set both weights to 1, though these can be adjusted depending on the operational focus.
Based on these definitions, the objective function can be represented as follows:

\begin{align}
    f(\cdot) &= w_{\text{mean}}f_{\text{mean}}(\cdot) + w_{\text{max}}f_{\text{max}}(\cdot) \label{eq:mainobj}\\
    f_{\text{mean}}(\cdot) &= \frac{1}{N_V} \sum_{k=1}^{N_V} OT_k \label{eq:meanobj}\\
    f_{\text{travel}}^k(X_k) &= \sum_{i=0}^{N_U} \sum_{j=0, j \neq i}^{N_U} x_{i,j}^k c_{i,j} \label{eq:travelobj} \\
    f_{\text{stay}}^k(X_k, S^+, S^-) &= \sum_{i=1}^{N_U}\sum_{j=1, j\neq i}^{N_U} x_{i,j}^{k}(t_j-(t_i+T_i^{\text{task}}+c_{i,j}) \nonumber \\ & +T_j^{\text{task}}) + \sum_{j=1}^{N_U} x_{0,j}^{k}T_{j}^{\text{task}}  \label{eq:stayobj}\\
    f_{\text{max}}(\cdot) &= z \label{eq:maxobj} 
\end{align}

Among these expressions, (\ref{eq:stayobj}) and (\ref{eq:maxobj}) are nonlinear and thus need linearization. To linearize (\ref{eq:stayobj}), we introduce two continuous auxiliary variables $s_{j,k}^+$ and $s_{j,k}^-$ to avoid a product of variables. They take $t_j$ and $-t_i$ respectively when $x_{i,j}^{k}=1$ and $0$ otherwise, which is enforced  using additional constraints. By using these auxiliary variables, the equation can be reformulated as below:
\begin{align}
    f_{\text{stay}}^k(X_k, S^+, S^-) &=\sum_{j=1}^{N_U} (s_{j,k}^+ + s_{j,k}^-) + \sum_{j=1}^{N_U} x_{0,j}^k T_j^{\text{task}} \nonumber \\
    &\quad + \sum_{i=1}^{N_U} \sum_{j=1}^{N_U} x_{i,j}^k (T_{j}^{\text{task}} - T_{i}^{\text{task}} - c_{i,j})
\end{align}
\begin{align}
    s_{j,k}^{+} &\leq M \sum\nolimits_{i=1,i\neq j}^{N_U}x_{i,j}^{k}+\epsilon \label{eq:sp1} \\
    s_{j,k}^{+} &\leq t_{j}+\epsilon +M (1-\sum\nolimits_{i=1,i\neq j}^{N_U}x_{i,j}^{k}) \label{eq:sp2} \\
    s_{j,k}^{+} &\geq t_{j}-\epsilon -M (1-\sum\nolimits_{i=1,i\neq j}^{N_U}x_{i,j}^{k}) \label{eq:sp3} \\
    s_{i,k}^{-} &\geq -M\sum\nolimits_{j=1, i\neq j}^{N_U}x_{i,j}^{k} - \epsilon \label{eq:sm1} \\
    s_{i,k}^{-} &\geq -t_{i} - \epsilon - M (1-\sum\nolimits_{j=1, i\neq j}^{N_U}x_{i,j}^{k}) \label{eq:sm2} \\
    s_{i,k}^{-} &\leq -t_{i} + \epsilon + M (1-\sum\nolimits_{j=1, i\neq j}^{N_U}x_{i,j}^{k}) \label{eq:sm3}
\end{align}

Furthermore, to represent MaxOT, we introduce an auxiliary variable $z$, and the maximum value is defined by the following constraint equation \cite{posner1981linear}. We could use the standard deviation instead of the maximum value, but we did not since it is difficult to incorporate as a linear objective.
\begin{align}
    (\forall k \in V) [OT_k \leq z] \label{eq:maxconst}
\end{align}
}

%% file: 04_algorithm.tex
\section{Method To solve nursing care taxi dispatch} \label{sec:Method}

\begin{algorithm}[t]
\caption{Nursing Care Taxi Dispatch Algorithm}
\label{alg:algo_all}
\begin{algorithmic}[1]
\renewcommand{\algorithmicrequire}{\textbf{Input:}}
\renewcommand{\algorithmicensure}{\textbf{Output:}}
\Require problem instances $s$
\Ensure trajectories $\pi'$
\While{True}
\State $\pi$ $\leftarrow$ AttentionModel$(s)$
\State $\pi'$ $\leftarrow$ post-process$(s, \pi)$
\If{$\pi'$ includes $N_V$ or less vehicles}
\State break
\EndIf
\EndWhile
\end{algorithmic}
\end{algorithm}
We propose a new algorithm that combines Attention Mechanism \cite{vaswani2017attention} and a heuristic algorithm to satisfy all the constraints (Algorithm \ref{alg:algo_all}).
\rev{
While conventional methods have successfully produced high-quality solutions for problems with relatively few constraints, such as TSP and CVRP, our problem setting involves numerous constraints related to user attributes and the number of vehicles, where constraint satisfaction depends on the specific values of these attributes. To address this challenge, we first employ supervised learning using
 the ILP solvers’ solutions to derive trajectory representations
 with low \revii{operating times}.
}
Moreover, we introduce two mechanisms to satisfy all the constraints in NCTD. In NCTD, we have two types of constraints: those whose violation can be checked only based on the next user, and those whose violation can be checked after visiting all users. For the former, we apply NCTD-specific masks based on the masks introduced in \cite{kool2018attention} (Section \ref{sec:masks}) during the architecture's inference. For the latter, we adopt the Insertion Algorithm \cite{rosenkrantz1977analysis} after obtaining a temporary solution using a Transformer-based approach (Section \ref{sec:ArrangingTours}).
\subsection{Transformer-based Architecture} \label{sec:am}
\subsubsection{Architecture} \label{sec:arch}
\begin{figure*}[!t]
    \centering
    \includegraphics[scale=.85]{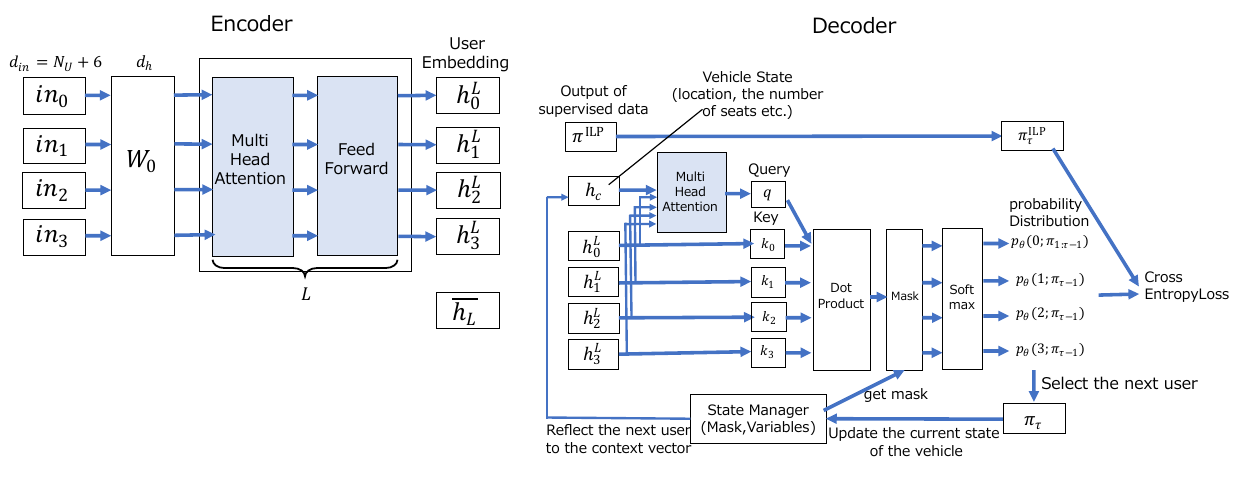}
    \caption{Machine Learning Architecture using Attention Mechanism}
    \label{fig:Architecture}
\end{figure*}
\rev{
Fig. \ref{fig:Architecture} shows our proposed architecture inspired by \cite{kool2018attention}. 
}
The architecture takes as input the attributes of the facility $in_0$ and all users $in_i(i\in U)$ and produces as output the trajectories of all vehicles $\pi=(\pi_{1},\dots,\pi_{n})(N_U+N_V\leq n \leq 2N_U)$, where $\pi_{k} (k=1,\dots,n)$ is the facility or one of the users, consisting of two parts: Encoder and Decoder.
\rev{
The encoder generates embedding vectors for each node $ i$ from the feature vector $in_i$:
\begin{align}
    in_{i} &= [x_i,y_i,wheel_i, T_i^S, T_i^G, T_i^{\text{task}}; inc_{i}]
\end{align}
First, the encoder produces the embedded vector $h^0_i$ whose dimension is $d_h$ from the input $in_i(\in \mathbb{R}^{d_{in}})$:
\begin{equation}
    h_{i}^{0} = W_0in_{i} + b \hspace{2mm} (\in \mathbb{R}^{d_h})
\end{equation}
where $W_0 \in \mathbb{R}^{d_h \times d_{in}}$ , $b \in \mathbb{R}^{d_h}$ are learnable parameters. 
Next, $h^0_i$ are passed through the Multi Head Attention (MHA) layer and the Feed Forward (FF) layer $L$ times to capture the relationship between users:
{
\begin{align}
    \widehat{h^l_i} & = \text{MHA}(h_0^l,\dots,h_{N_U}^l) + h_i^l  \\
    h_i^{l+1} & = \text{FF}(\widehat{h_i^l}) + \widehat{h_i^l}
\end{align}
Here, MHA is calculated as follows based on \cite{vaswani2017attention}. The query, key, and value are computed from the input matrix, and the embedding dimension is split into $\eta$ heads. Attention is then calculated for each head, and the outputs are subsequently concatenated to form the final MHA:
\begin{align}
    H &= [h_0^l;\dots;h_{N_U}^l]^{\top} 
\end{align}
\begin{align}
    Q=HW^Q, K=HW^K, V=HW^V \\
    Q_i=QW_{i}^{Q}, K_i=KW_i^K, V_i=VW_i^V
\end{align}
\begin{align}    
    \text{head}_i &= \text{softmax}(\frac{Q_iK_i^{\top}}{\sqrt{d_k}})V_i \\
    \text{MHA}(H) &= \text{concat}(\text{head}_1, \dots, \text{head}_{\eta})W^O
\end{align}
where each head's dimension of keys and values is $d_k=\frac{d_h}{\eta}$, learnable parameters are matrices $W^Q, W^K, W^V\in \mathbb{R}^{d_h\times d_h}$, $W_i^Q, W_i^K, W_i^V\in \mathbb{R}^{d_h\times d_k}$, and $W^O\in \mathbb{R}^{d_h\times d_h}$.
}
Finally, the encoder outputs a user embedding vector $h^L_i$, which contains the linkage information between users, and a graph embedding vector $\overline{h^L}$ obtained by averaging the user embedding vectors.

The Decoder, whose structure is based on \cite{kool2018attention},  receives user embedding vectors from the Encoder’s output, along with a context vector $h_c$ that reflects the vehicle’s state at step $\tau$:
\begin{equation}
    h_c = [\overline{h^L};h_{\pi_{\tau-1}},R_{\tau}^W, R_{\tau}^N, curtime_{\tau}, vc_{\tau}^1, vc_{\tau}^2;  inc_{\tau}^{\text{board}}]
\end{equation}
First, a context vector $h_c$ is passed through an MHA layer, obtaining the query vector $q$. This attention step is called \textit{Glimpse}\cite{bello2016neural}, allowing \revii{the learning model} to emphasize information of nodes that are highly relevant to the current vehicle state:
\begin{align}
    h_c^{'} &= \text{FF}(h_c)\\
    H &= [h_{0}^L; \dots; h_{N_U}^L]^{\top} \\
    q &= \text{MHA}(h_c^{'}W_Q, HW_K ,HW_V) \\
    K&=HW_K 
\end{align}

Subsequently, the dot product is calculated between $q$ and the node embeddings from the Encoder. The result is then passed through a tanh function and scaled by a constant $C$ to obtain logits $u_{(c)j}$ within the range $[-C,C]$. A softmax function is applied to these logits to obtain the visit probabilities $p_j$. When calculating $u_{(c)j}$, the mask is utilized to identify constraint violations. If user $j$'s mask indicates true, then the probability is 0 based on (\ref{eq:ucj}).
\begin{align}
    u_{(c)} &= \begin{cases}
        C\tanh (\frac{qK^{\top}}{\sqrt{d_h}}) \\
        -\infty
    \end{cases} \label{eq:ucj} \\ 
    u_{(c)j}& = [u_{(c)}]_j \\
    p_j &= \text{softmax}(u_{(c)})_j
\end{align}
}

\subsubsection{Applying Constraints} \label{sec:masks}
\begin{table}[t]
    \centering
    \caption{Variables kept in our \revii{learning model}}
    \begin{tabular}{ll}
    \hline
        \textbf{Notation} & \textbf{Description} \\
        \hline
        $user^W_{\tau}$ & the number of wheelchair users at step $\tau$ \\ 
        $user^N_{\tau}$ & the number of normal users at step $\tau$  \\
        $inc_{\tau}^{\text{board}}$ & the logical sum of passengers' incompatibility at step $\tau$\\
        $T^G_{min,\tau}$ &the minimum $T_j^G$ of all users on board \\
        $curtime_\tau$ & the time at the step $\tau$ \\
        $vc_{\tau}^1$ & the number of remaining vehicles at step $\tau$\\
        $vc_{\tau}^2$ & an indicator of vehicle sufficiency at step $\tau$\\
        \hline
    \end{tabular}
    \label{tab:notation_table_val_ml}
\end{table}
Kool et al. \cite{kool2018attention} introduced some variables that represent the vehicle's state at step $\tau$, which are used to check constraint violations and update the context vector. Based on this, we define seven variables shown in Table \ref{tab:notation_table_val_ml} and update them by (\ref{eq:user_w})-(\ref{eq:vehicleConst2}):
\begin{align}
    user^W_{\tau+1} &\leftarrow user^W_{\tau} + \frac{wheel_{j}}{cap_{W}} \label{eq:user_w} \\
    user^N_{\tau+1} &\leftarrow user^N_{\tau} + \frac{1-wheel_{j}}{cap_{N}} \label{eq:user_n}\\
    inc_{\tau+1}^{\text{board}} &\leftarrow inc_{\tau}^{\text{board}} \vee inc_{j} \label{eq:prf}\\
    T^G_{min,\tau+1} &\leftarrow \text{min}(T^G_{min,\tau}, T^G_{j}) \label{eq:tgmin}\\ 
    curtime_{\tau+1} &\leftarrow \text{max}(curtime_{\tau}+T_{i}^{\text{task}}+c_{i,j}, T^S_j - \alpha) \label{eq:curtime} \\
    vc_{\tau}^{1} &= \frac{Rem_{V}}{N_V}  \label{eq:vehicleConst} \\
    vc_{\tau}^2 &= \tanh(1- \frac{Rem_{U}N_{V}}{Rem_{V}N_{U}}) \label{eq:vehicleConst2} 
\end{align}
Where $cap_{W}, cap_{N}$ are the capacity of wheelchair seats and normal seats at step $\tau$, \rev{ and $Rem_U$ is the number of unvisited customers and $Rem_V$ is the number of remaining vehicles at step $\tau$.}
When user $j$ is picked up at time step $\tau$, $user_{\tau}^W$ is updated by (\ref{eq:user_w}) if the user requires a wheelchair; otherwise, $user_{\tau}^N$ is updated by (\ref{eq:user_n}). \rev{ Moreover, user $j$’s incompatibility is reflected to the current vehicle by updating $inc_{\tau}^{\text{board}}$ by (\ref{eq:prf}). By calculating the logical sum, if at least one passenger has an incompatibility with  user $i$, the $i$-th component of $inc_{\tau}^{\text{board}}$ is set to 1, which can prevent user $i$ from riding.} $T_{min,\tau}^{G}$ represents the earliest arrival time at the facility among all users on board at time step $\tau$, ensuring that all people will not be late. It is updated to $+\infty$ only when the vehicle visits the facility since no new users will be assigned. $curtime_{\tau}$ means the current time. This is updated by adding the time for tasks at the user's house $T_{i}^{\text{task}}$ and the travel time $c_{i,j}$, but the vehicle waits until $T_i^S - \alpha$ when reaching the house before the preferred time slot. 
\rev{
$vc_{\tau}^1, vc_{\tau}^2$ are variables related to vehicle number constraint. The value $vc_{\tau}^1$ corresponds to the remaining number of vehicles. A smaller value indicates fewer remaining vehicles, implying a higher risk of violating the vehicle-number constraint. For $ vc_{\tau}^2$, the idea is based on comparing the average number of passengers per vehicle at the beginning of the routing and after $\tau$ steps of visiting. Initially, the average capacity is $N_U/N_V$, and at step $\tau$, the average capacity associated with the remaining unvisited customers can be represented as $Rem_U/Rem_V$. When $Rem_U/Rem_V$ is smaller, the current routing has more spare vehicles. In such cases, the following inequality holds:
\begin{equation}
    \frac{N_U}{N_V}>\frac{Rem_{U}}{Rem_{V}} \Leftrightarrow 1-\frac{Rem_{U}N_V}{Rem_{V}N_U} > 0
\end{equation}
A positive value of $ vc_{\tau}^2$ therefore indicates that sufficient vehicles remain, whereas negative values imply that a vehicle-number violation is more likely to occur. 
}
\begin{align}
    mask_{i}^{1} & = (user^W_{\tau} + \frac{wheel_{j}}{cap_{W}} > 1) \label{eq:ML_wheelconstraint} \\
    mask_{i}^{2} & = (user^N_{\tau} + \frac{1-wheel_{j}}{cap_{N}} > 1) \label{eq:ML_normalconstraint} \\
    mask_{i}^{3} & = (inc_{\tau}^{\text{board}} = 1 )\label{eq:pref}\\
    mask_{i}^{4} & = (curtime_{\tau} + T_{i}^{\text{task}} + c_{i,j}  >  T^S_j + \alpha) \label{eq:time_upper} \\
    mask_{i}^{5} & = (T^S_j + \alpha + c_{j,0} > T^G_{min,\tau} - \beta) \label{eq:arrival} \\
    mask_{i,\tau} & = \bigvee\nolimits_{m=1}^{5}mask_{i}^{m} \label{eq:maskall}
\end{align}
We judge the constraint violations using flags (\ref{eq:ML_wheelconstraint})-(\ref{eq:arrival}), which correspond to constraints described in Section \ref{sec:Formulation}. Equation (\ref{eq:maskall}) is the logical sum of all masks, whose value will be 1 if at least one constraint violation occurs. 

\subsubsection{Training} \label{sec:training}
\begin{algorithm}[t]
    \caption{Training}
    \label{alg:algo_learning}
    \begin{algorithmic}[1]
        \renewcommand{\algorithmicrequire}{\textbf{Input:}}
        \renewcommand{\algorithmicensure}{\textbf{Output:}}
        \Require Epoch $E$, the number of steps $S$, the batch size $B$, supervised data $(s_{i}, \pi^{\text{ILP}(i)}) (i={1,..,N_d})$ 
        \For{$e=1,..,E$}
        \For{$s=1,\dots,S$}
        \State $Loss = 0$
        \For{$b=1,\dots,B$}
        \For{$t=1,...,T$}
        \State Calculate the probability distribution $p_{\theta}^{\text{prob}}$
        \State Obtain the ground truth
        \State Update $loss_{b,\tau}$ by (\ref{eq:partof_celoss})
        \State $Loss \leftarrow Loss + loss_{b,\tau}$
        \State Let the vehicle move to the user $\pi_{\tau}^{\text{ILP}(b)}$
        \EndFor
        \EndFor
        \State Divide $Loss$ by $BT$
        \State Calculate $\nabla Loss$ 
        \State $\theta \leftarrow \text{Adam}(\theta,\nabla Loss)$
        \EndFor
        \EndFor
    \end{algorithmic}
\end{algorithm}
High-quality solutions can be obtained by applying the formulation to the ILP solver. The solution is described as sequential data of visited users: $\pi^{\text{ILP}}=(\pi_{1}^{\text{ILP}},\dots,\pi_{T}^{\text{ILP}})$, where $T$ is the length of the sequence, which is up to $2N_{U}$. We use supervised learning to decrease the difference between two sequences: a solution generated by machine learning and the ground truth generated by the ILP solver (Algorithm \ref{alg:algo_learning}). We apply the cross-entropy loss as a loss function, which is calculated by the following equation, using each component of the ground truth and the probability $p_{\theta(j)}$ produced from the decoder:

\begin{align}
    \text{CELoss}&= \frac{1}{BT}\sum\nolimits_{b=1}^{B}\sum\nolimits_{\tau=1}^{T}loss_{b,\tau} \label{eq:celoss} \\
    loss_{b,\tau} &= \sum\nolimits_{j=0}^{N_U}p_{b,\tau,j}^{\text{label}}\log p_{\theta}(j|\pi_{b,1:\tau-1},s_{b}) \label{eq:partof_celoss}
\end{align}

where $B$ is the batch size, $T$ is the total number of steps, and $p_{b,\tau,j}^{\text{label}}$ is the $j$-th label of the ground truth at step \revii{$\tau$} of instance $b$, taking 1 when $j=\pi_{\tau}^{\text{ILP}(b)}$ and 0 otherwise.

While the next user is determined based on the probability distribution in the inference phase, the distribution is used only to compute the loss function during training, and vehicles forcefully visit the ground truth user at each step.
Our decision to strictly follow the ground truth path when training is motivated by the observation that the probability distribution is conditioned on the previous path, particularly the immediately preceding user. This is because, once the vehicle deviates from the ground truth path, there is a risk of generating a sequence that diverges from the ground truth trajectory, potentially hindering effective learning.
\subsubsection{Training Data} \label{sec:traindata}
The input data for a single problem instance includes values shown in Table \ref{tab:notation_table}. Values related to location ($loc_0, loc_i$ where $i=1, ..., N_U$) and time ($T_i^S, T_i^G, T_i^{\text{task}}$) vary across regions. Therefore, we normalize these values to the range of $[0,1]$ to eliminate differences in values of the locations and unify the data scale.
\rev{
The specific inputs are structured as follows. The location data of the facility and users are represented as a list of (x, y) coordinates. The wheelchair usage status is given by a list of length $N_U$, where each element is denoted by $wheel_i(i=1,\dots,N_U)$. User incompatibility is input as a matrix with binary values (0 or 1). The scheduled pickup times, arrival times at the facilities, and the task durations are respectively input as lists of normalized values. Other inputs include scalar values, such as the parameters shown in Table \ref{tab:timeparam}, the region range (in kilometers), and a time upper bound (the end time of the scheduling period). 
}
The training data consists of two types: real data and \revii{uniformly distributed synthetic data (synthetic data for short).}
The real data utilize the location and expected departure time from the real-world dataset described in Section \ref{sec:Datasets}, \revii{while the synthetic data are generated entirely based on a uniform distribution.}
\revii{The synthetic data are not intended to approximate or replace real-world data, but are introduced as a structure-free baseline without any specific geographical characteristics.}
\revii{By preparing both types of data, we investigate how the presence or absence of real-world geographical structure affects the solution quality. In addition, the use of synthetic data allows us to discuss the method without involving sensitive user information.}

\subsection{Post-Processing of the Generated Tours} \label{sec:ArrangingTours}
Since the total number of vehicles cannot be determined only using masks described in Section \ref{sec:am}, which can only detect the constraint violations occurring one step ahead, we need an additional mechanism that allows the solution to be feasible.
We incorporate the heuristic algorithm for TSP, the Insertion Algorithm \cite{rosenkrantz1977analysis}, and modify the outputs of the \revii{learned model} to obtain feasible solutions (Algorithm \ref{alg:algo_arrange}). The main procedure consists of two steps: (1) Deleting paths and (2) Inserting users.

\subsubsection{Deleting Paths}
First, we divide the solution into individual routes for each vehicle $\rho_i(i=1,\dots, N_{V}^{'})$ to determine the number of vehicles used. 
If the number of vehicles $N_{V}^{'}$ is less than or equal to $N_V$, we consider this the final solution since that trajectory is feasible. On the other hand, if more than $N_V$ vehicles are used, we remove $N_{V}' - N_V$ redundant routes to reduce the total number of vehicles to $N_V$. To minimize the users included in the removed routes, we sort the routes by the number of passengers in ascending order and then remove the top $N_{V}' - N_V$ routes, assuming that we remove $r$ users ($\nu_1,\dots,\nu_r$). Then we get a temporary solution denoted as $\pi^{'}$, which meets Vehicle Number Constraint but does not include all users.
\subsubsection{Inserting Users}
\begin{algorithm}[t]
    \caption{Post-Processing Algorithm}
    \label{alg:algo_arrange}
    \begin{algorithmic}[1]
        \renewcommand{\algorithmicrequire}{\textbf{Input:}}
        \renewcommand{\algorithmicensure}{\textbf{Output:}}
        \Require problem instance $s$, tentative solution $\pi$, the available number of vehicles $N_V$
        \Ensure modified solution $\pi'$
        \State separate $\pi$ to each vehicle's trajectory $\rho_{i}(i=1,\dots,N'_{V})$
        \If{$N'_{V} \leq N_{V}$}
        \State \Return{$\pi$}
        \Else
        \State AscendingSort($\rho_{1},\dots,\rho_{N'_{V}}$)
        \State $\nu_{1},\dots,\nu_{r} \leftarrow \text{removeRoute}(\rho_{1},\dots,\rho_{N'_{V}})$
        \State $\pi' \leftarrow \text{concat}(\rho_{N'_{V}-N_{V}+1},\dots,\rho_{N'_{V}})$ 
        \For{$i=1,\dots,r$}
        \State $cost_{\text{curr}} = \infty, user_{\text{ins}} = -1$ 
        \For{$j=1,\dots,\text{len}(\pi')-1$}
        \State insert $\nu_{i}$ between $\pi'_{j}$ and $\pi'_{j+1}$
        \State check constraints(Section \ref{sec:capConst},\ref{sec:timeConst},\ref{sec:Prefconst})
        \State $cost_{\text{new}} = c_{\pi'_{j},\nu_{i}} + c_{\nu_{i},\pi'_{j+1}} - c_{\pi'_{j},\pi'_{j+1}}$
        \If{satisfy constraints}
        \If{$cost_{\text{new}} < cost_{\text{curr}}$}
        \State $cost_{\text{curr}} = cost_{\text{new}}$
        \State $user_{\text{ins}} = j$
        \EndIf
        \Else
        \State remove $\nu_i$ between $\pi'_{j}$ and $\pi'_{j+1}$
        \EndIf
        \EndFor
        \If{$user_{\text{ins}} == -1$}
        \State \Return{no solution}
        \EndIf
        \EndFor
        \State \Return{$\pi'$}
        \EndIf
    \end{algorithmic}
\end{algorithm}
%
As a result of removing some routes, we inevitably have unvisited users $\nu_1,\dots,\nu_r$. To obtain a high-quality feasible solution, we refer to the Insertion Algorithm \cite{rosenkrantz1977analysis}, a heuristic algorithm to solve the TSP, \rev{ and consider reinserting $r$ unassigned users into existing tours $\pi^{'}$, performed one user at a time. }
\rev{ For reinsertion of each node, all edges between adjacent nodes within the tour $\pi'$ become candidates for insertion and we determine whether it satisfies three types of constraints (Capacity Constraint, Time Constraint, and Incompatibility Constraint presented in Section \ref{sec:Formulation}) for each candidate.}
If we find a position satisfying all three constraints, we calculate a new cost increase, denoted as $cost_{\text{new}}$.
To prevent an excessive increase in the \revii{operating time}, we greedily search for the insertion points where the increase in operating time is minimized. If this calculated cost increase is smaller than the current least cost increase $cost_{\text{curr}}$, we update $cost_{\text{curr}}$.
If all users can be inserted while satisfying all constraints, we consider this the final solution. However, there may be cases where no position can be found to insert a user. \rev{In such cases, since the solution is infeasible, the algorithm returns nothing. We must then recompute the solution from scratch, as described in Algorithm \ref{alg:algo_all}.}

\subsection{Analyses of Post Processing}
\rev{
\subsubsection{Edge Cases of Post-Processing}
When the number of available vehicles is low, the initial solution generated by the \revii{learning model} often exceeds the predefined maximum number of available vehicles. Consequently, a larger number of nodes must be reinserted into the tentative solution during post-processing. This, in turn, lowers the probability of assigning all unvisited nodes while satisfying all the mask-based constraints, making it more likely that the final obtained solution will be infeasible.
In conclusion, an insufficient supply of vehicles complicates post-processing adjustments, increasing the risk of obtaining a solution that violates the vehicle number constraint. In contrast, mask-based constraints, such as user compatibility and pickup time, are always satisfied at the \revii{learning model’s} mask mechanism during inference, so these constraints will not be violated in the post-processing stage.
}

\rev{
\subsubsection{Time Complexity of Post Processing}
First, in the path deleting phase, the process of partitioning the tentative solution $\pi$ into vehicle routes requires $O(\text{len}(\pi)) = O(N_U)$ computational complexity because this process scans $\pi$ linearly and the length of $\pi$ is up to $2N_U$, while the process of sorting each route in descending order by visit count requires $O(N_V \log{N_V})$ computational complexity, if we use a quick sorting algorithm such as quicksort.
Next, in the reinsertion phase, $r$ can be as large as $N_U - N_V$
 in the worst case, but is typically smaller practically because it is expected that the learning model will capture a good representation of the trajectory. For each node insertion, we attempt insertion between two adjacent node in the tour (worst-case length is $O(N_U)$)
 and evaluate the constraints. The time complexities of Capacity Constraint, Time Constraint, and Incompatibility Constraint are $O(\text{len}(\rho))$, $O(\text{len}(\rho)^2)$, and $O(N_U)$ respectively, where $\rho$ represents a trajectory partitioned by each vehicle. Since the tour lengths of vehicles are constrained by the maximum seat capacity $M_{cap} = \max_{k}(cap_k^W + cap_k^N)$, the computational complexity of this decision process is $O(M_{cap}^2)$, where the time constraint dominates.
 From these analysis, the overall computational complexity of post-processing is $O(N_U + N_V\log N_V + N_U(N_U-N_V)M_{cap}^2)$ in the worst case.
 }

%% file: 05_evalation.tex
\section{Evaluation} \label{sec:Evalation}
\subsection{Datasets and Evaluation Data} \label{sec:Datasets}
\begin{table}[t]
    \centering
    \caption{The range of user attribute values and available vehicles}
    \begin{tabular}{c|c|c|c}
        \hline
        Users & \begin{tabular}{c}Wheelchair\\Users\end{tabular} & \begin{tabular}{c}the number of\\incompatibility pairs\end{tabular} & \begin{tabular}{c}the available \\number of cars\end{tabular} \\ \hline
        20 & 1-4 & 1-4 & 5 \\
        30 & 1-6 & 1-6 & 7\\ 
        50 & 1-10 & 1-10 & 12\\ \hline
    \end{tabular}
    \label{tab:userattr}
\end{table}
\begin{figure}[t]
    \centering
    \includegraphics[width=.9\linewidth]{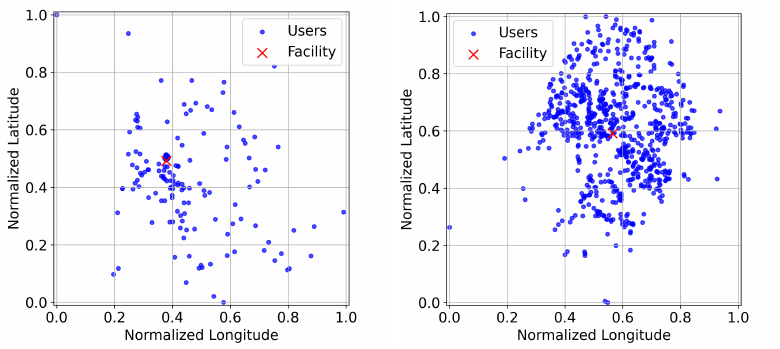}
    \caption{\rev{ Users' locational distributions of Gunma Facility(left) and Osaka Facility(right)}}
    \label{fig:loc_distl}
\end{figure}
\begin{figure}[t]
    \centering
    \includegraphics[width=.9\linewidth]{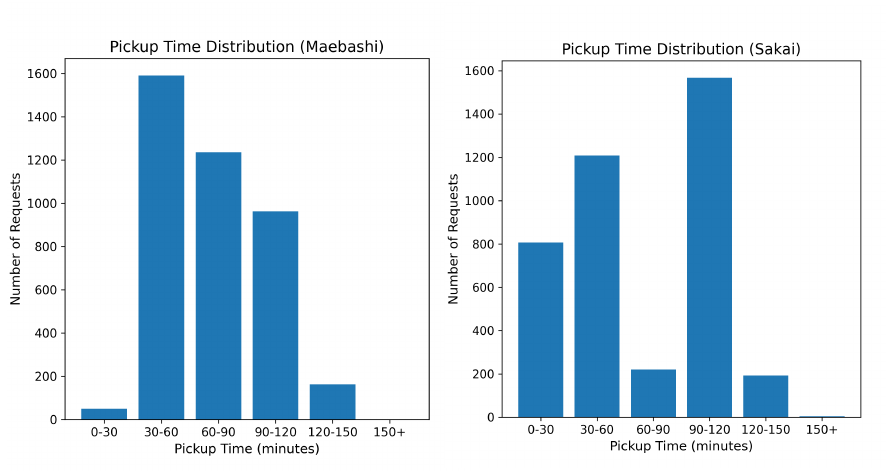}
    \caption{\rev{ Temporal distributions of users' pickup time of Gunma Facility(left) and Osaka Facility(right)}}
    \label{fig:temp_distl}
\end{figure}
We utilize care taxi transportation history data collected by the JST CREST program\footnote{https://www.jst.go.jp/kisoken/crest/en/} as our real-world data. These data
are obtained from different facilities in two different cities in Japan: facilities in Maebashi City, Gunma Prefecture (simply \textit{Gunma Facility}) and in Sakai City, Osaka Prefecture (simply \textit{Osaka Facility}). We utilize transportation data recorded over 76 days from January to March 2023 at the Gunma Facility and 31 days in October 2023 at the Osaka Facility. 
\rev{ We show users' locational distributions of two facilities in Fig. \ref{fig:loc_distl}. In the Gunma Facility’s dataset, users are distributed around the facility, with a relatively large number of users clustered on the east and south sides of the facility. Users are distributed over a range of approximately 12.3 km from east to west, and approximately 12.1 km from north to south, with a total of 153 users. Similarly, in the Sakai City’s dataset, users are concentrically distributed around the facility, but there are some areas where few users can be seen due to some large parks. The east-west and north-south range are approximately 18.5 km and 19.9 km, respectively, and a total of 730 users are using the facility. As for the distribution of pickup request times, the distribution was created in 30-minute segments based on the request times taken from the test data in Fig. \ref{fig:temp_distl}. As well as the distribution of location, the distribution of request times also differs between two cities. For example, the distribution of requests is over a wider time period in Sakai, and different peaks can be observed.}

We employ this real-world data to generate problem instances (datasets) for the NCTD. When creating problem instances of real data, we extract the locations of houses and expected departure times from the real data to reflect actual management. We then randomly set user attributes within the range shown in Table \ref{tab:userattr} to account for the diversity of user types.

\subsection{Data Modification}\label{sec:DataModi}
\begin{figure}
    \centering
    \includegraphics[width=1\linewidth]{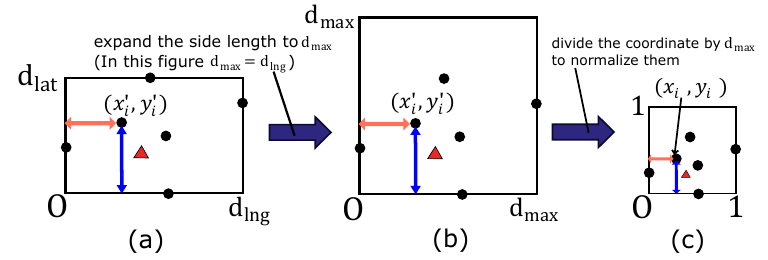}
    \caption{ The translation of the latitude and longitude into the normalized coordinate}
    \label{fig:dataModi}
\end{figure}
We need to transform each user's longitude and latitude $(\text{lng}_i,\text{lat}_i)$ to normalized coordinates $(x_i, y_i)$ in the range of $[0,1]$, as described in Section \ref{sec:traindata}. First, we decide the base point as ($\text{min}_{\text{lat}}$,$\text{min}_{\text{lng}}$), which are the smallest latitude and longitude in the given problem instance. We also define ($\text{max}_{\text{lat}}$,$\text{max}_{\text{lng}}$) as the largest latitude and longitude in the given problem instance.  Then, we calculate for each user the east-west distance from the base point (denoted as $x_{i}'$) and the north-south distance from the base point (denoted as $y_{i}'$) (Fig. \ref{fig:dataModi}(a)). We use Karney's algorithm \cite{karney2013algorithms} when calculating distances:
\begin{align}
    x_{i}' &= \text{dist}((\text{lat}_i, \text{lng}_i), (\text{lat}_i, \text{min}_{\text{lng}})) \\
    y_{i}' &= \text{dist}((\text{lat}_i, \text{lng}_i),(\text{min}_{\text{lat}}, \text{lng}_i))
\end{align}
Next, we calculate the range in an east-west direction $d_{\text{lat}}$ and the range in a north-south direction $d_{\text{lng}}$, and define $d_{\text{max}}$ as the bigger of $d_{\text{lat}}$ and $d_{\text{lng}}$ (Fig. \ref{fig:dataModi}(b)):
\begin{align}
    d_{\text{lng}} &= \text{dist}((\text{min}_{\text{lat}}, \text{min}_{\text{lng}}), (\text{min}_{\text{lat}}, \text{max}_{\text{lng}})) \\
    d_{\text{lat}} &= \text{dist}((\text{min}_{\text{lat}}, \text{min}_{\text{lng}}), (\text{max}_{\text{lat}}, \text{min}_{\text{lng}})) \\
    d_{\text{max}} &= \max(d_{\text{lng}}, d_{\text{lat}})
\end{align}
 Finally, we divide $x_{i}'$, $y_{i}'$ by $d_{\text{max}}$ to obtain the normalized coordinate $x_{i}$, $y_{i}$ (Fig. \ref{fig:dataModi}(c)) :
\begin{align}
    x_i &= x_{i}' / d_{\text{max}} \\
    y_i &= y_{i}' / d_{\text{max}} 
\end{align}
Strictly speaking, the distance of one degree of longitude differs at different latitudes, but since the covered area is small in this problem (about 10-20 kilometers per side), we do not have the above issue in this research.

\subsection{Evaluation Method}\label{sec:EvalMethod}
\rev{
\begin{table}[t]
    \centering
    \caption{The number of training and test datasets}
    \begin{tabular}{c|c|c|c}
        \hline
         & $N_U=20$ & $N_U=30$ & $N_U=50$ \\ \hline
         training (synthetic dataset) & 2592 & 2784 & 1792 \\
         training (real dataset) & 5760 & 5728 & 2528 \\ 
         test dataset (Gunma Facility) & 100 & 99 & 100 \\ 
         test dataset (Osaka Facility) & 100 & 100 & 100 \\ \hline
    \end{tabular}
    \label{tab:data_table}
\end{table}
}
To simulate various care facilities with different numbers of users, we set the number of users $N_U$ to 20, 30, and 50. For each user number, the numbers of training and test datasets are set as shown in Table \ref{tab:data_table}.
The training datasets consist of two types as described in Section \ref{sec:traindata}: real datasets generated from the real-world data introduced in Section \ref{sec:Datasets}, and \revii{synthetic datasets based on a uniform distribution}. The test datasets are entirely generated from real-world data.

\rev{As hyperparameters, we have the Encoder's embedding dimension $d_h$, the MHA head $\eta$, and the number of layers $L$. To find the optimal embedding dimension for our input dimension, we conducted experiments by varying $d_h$ to $64, 128, 256$. For the MHA head, as our training data comprises only a few thousand samples, we set $\eta=8$. Furthermore, we set the number of layers to $L=3$, as recommended in the previous paper \cite{kool2018attention}.}

We compared our proposed method with three existing methods using neural networks: AM\cite{kool2018attention}, Policy Optimization with Multiple Optima (POMO) \cite{kwon2020pomo}, and Adaptive Multi-Distribution Knowledge Distillation Attention Model (AMDKD-AM) \cite{bi2022learning}\revii{, and two GA-based algorithms: NSGA-II \cite{deb2002fast} and MOEA/D\cite{zhang2007moea}.}  \rev{Since \revii{ML-based }baselines for comparison \revii{are methods based on end-to-end learning models} trained in a learning-based manner, the solutions obtained in different executions are not guaranteed to be identical due to stochasticity of the method, which may lead to an unreliable evaluation. To improve reproducibility, we execute \revii{each method} five times under identical settings, and record the best \revii{operating time} obtained for each instance. This allows for a more equitable comparison of the performance of the different methods. To compute the ground truth, we used an ILP solver (IBM CPLEX 22.1.1 and HiGHS 1.11.0). For $N_U=20,30$, we used 60-second results by CPLEX for training to ensure a sufficient dataset size. For $N_U=50$, to ensure the quality of ground truth, we first obtained solutions by HiGHS for 21000 seconds, and then improved them by CPLEX for 300 seconds using them as initial solutions.
For AM and AMDKD-AM, to decrease the violation of vehicle number constraint, we added the penalty term to the objective function (\ref{eq:mainobj}), which increases the function's value when the violation of vehicle number constraint occurs, since it has no mechanism to suppress that constraint. Regarding POMO, the penalty term is not included due to the difficulty of counting the number of vehicles during implementation.}

\revii{For NSGA-II and MOEA/D, to incorporate our NCTD, constraint violations and objective function values were evaluated by Deb's feasibility rule\cite{deb2000efficient} when comparing two individuals. Specifically, if both violated constraints, the one with the smaller number of violated constraints was preferred; if only one violated constraints, the feasible one was preferred; and if both were feasible, solution quality was evaluated based on the objective function values. In NSGA-II, we use them for the rank obtained by non-dominated sorting, and in MOEA/D, we evaluate the solutions by a scalarizing function based on the Tchebycheff function\cite{zhang2007moea}. Both NSGA-II and MOEA/D used a population size of 200.
NSGA-II was executed for 400 generations, while MOEA/D was allowed 80,000 function evaluations, resulting in a comparable computational budget for both methods.
}

\rev{Furthermore, to consider various available times, we executed the ILP solver with different time limits. Limits under 10 minutes simulate situations where recomputation is required due to sudden changes in users' demands, such as adding new users or lacking users due to illness. In contrast, limits longer than 30 minutes simulates a situation where sufficient computation time is available.

The evaluation metrics include the average MeanOT and MaxOT (detailed in Section \ref{sec:Formulation}) of all feasible solutions, the sum of MeanOT and MaxOT (\revii{the operating time, OT} for short), \revii{execution time per instance}, and the rate of constraint violation \%(Viol.\% for short), which considers how many problem instances violate the vehicle number constraint among all instances.
To make the evaluation values easier to understand, we convert MeanOT and MaxOT from normalized scales to the original scales, allowing for comparison in minutes.
 We trained our \revii{learning model} on a single NVIDIA GeForce RTX 5060 Ti with a batch size of 32 and 100 epochs. We generated our training and test data using CPLEX running on AMD Ryzen 9 9950X 16-Core Processor for $N_U=20,30$, and HiGHS\cite{huangfu2018parallelizing} running on Supercomputer Fugaku with Fujitsu A64FX Processor for $N_U=50$.} 

\subsection{Results}\label{sec:results}

\begin{table*}[!t] 
\centering 
\caption{Comparison of Each Metrics Across Different Methods} 
\label{tab:results} 
\begin{tabularx}{\textwidth}{c|l|*{5}{X}|*{5}{X}} 
\hline
\multirow{2}{*}{Users} & \multirow{2}{*}{Method} & \multicolumn{5}{c|}{Gunma Facility} & \multicolumn{5}{c}{\revii{Osaka Facility}} \\ \cline{3-12}
& & MeanOT [min.] & MaxOT [min.] & OT [min.] & \revii{Execution Time} & Viol. \% & MeanOT [min.] & MaxOT [min.] & OT [min.] & \revii{Execution Time} & Viol. \%  \\ \hline
\multirow{11}{*}{20} & CPLEX(60s) & 14.332 & 20.489 & 34.821 & 60s & 0\% & 21.815 & 26.212 & 48.027 & 60s & 0\% \\
 & CPLEX(300s) & 14.406 & 20.221 & 34.626 & 300s & 0\% & 21.768 & 26.125 & 47.893 & 300s & 0\% \\
 & CPLEX(600s) & 14.318 & 20.277 & 34.595 & 600s & 0\% & 21.742 & 26.133 & 47.875 & 600s & 0\% \\
 & CPLEX(1800s) & 14.348 & 20.226 & 34.574 & 1800s & 0\% & 21.785 & 26.073 & 47.858 & 1800s & 0\% \\ \cline{2-12}
 & \revii{NSGA-II\cite{deb2002fast}} & 16.951 & 20.463 & 37.414 & 265.03s  & 0\% & 23.689 & 27.852 & 51.541 & 254.13s & 2\% \\
 & \revii{MOEA/D\cite{zhang2007moea}} & - & - & - & - & 100\% & - & - & - & - & 100\% \\ \cline{2-12}
 & AM\cite{kool2018attention}, real & 16.328 & 27.685 & 44.013 & 870ms & 1\% & 24.653 & 33.059 & 57.712 & 962ms & 10\% \\
 & POMO\cite{kwon2020pomo}, real & 14.610 & 22.516 & 37.127 & 421ms & 95\% & 22.174 & 29.901 & 52.076 & 435ms & 99\% \\
 & AMDKD-AM\cite{bi2022learning} & 15.079 & 26.677 & 41.756 & 1005ms & 2\% & 23.725 & 32.808 & 56.533 & 867ms & 10\% \\
 & \textbf{Ours, real} & \textbf{15.521} & \textbf{22.811} & \textbf{38.332} & \textbf{1778ms} & \textbf{0\%} & \textbf{23.265} & \textbf{28.987} & \textbf{52.252} & \textbf{2356ms} & \textbf{1\%} \\ \cline{2-12}
 & AM\cite{kool2018attention}, \revii{synthetic} & 18.741 & 27.929 & 46.670 & 875ms & 0\% & 26.259 & 36.198 & 62.456 & 838ms & 0\% \\
 & POMO\cite{kwon2020pomo}, \revii{synthetic} & 13.200 & 24.046 & 37.246 & 460ms & 98\% & - & - & - & - & 100\% \\
 & \textbf{Ours, \revii{synthetic}} & \textbf{16.777} & \textbf{24.017} & \textbf{40.794} & \textbf{1653ms} & \textbf{0\%} & \textbf{24.778} & \textbf{30.981} & \textbf{55.759} & \textbf{2622ms} & \textbf{0\%} \\ \hline\hline
\multirow{11}{*}{30} & CPLEX(60s) & 15.811 & 24.554 & 40.365 & 60s & 0\% & 22.139 & 29.151 & 51.290 & 60s & 0\% \\
 & CPLEX(300s) & 13.703 & 20.966 & 34.670 & 300s & 0\% & 20.532 & 25.838 & 46.370 & 300s & 0\% \\
 & CPLEX(600s) & 13.598 & 20.755 & 34.353 & 600s & 0\% & 20.476 & 25.716 & 46.192 & 600s & 0\% \\
 & CPLEX(1800s) & 13.421 & 20.649 & 34.070 & 1800s & 0\% & 20.433 & 25.476 & 45.909 & 1800s & 0\% \\ \cline{2-12}
  & \revii{NSGA-II\cite{deb2002fast}} & 17.411 & 21.224 & 38.635 & 679.09s & 1.01\%  & 26.471  & 30.803 & 57.273 & 651.08s & 7\% \\
 & \revii{MOEA/D\cite{zhang2007moea}} & - & - & - & - & 100\% & - & - & - & - & 100\% \\ \cline{2-12}
 & AM\cite{kool2018attention}, real & 16.137 & 28.473 & 44.610 & 1327ms & 0\% & 23.822 & 33.030 & 56.852 & 1416ms & 3\% \\
 & POMO\cite{kwon2020pomo}, real & 13.585 & 24.221 & 37.806 & 932ms & 88.89\% & 20.258 & 25.445 & 45.703 & 808ms & 96\% \\
 & AMDKD-AM\cite{bi2022learning} & 17.458 & 27.262 & 44.720 & 1183ms & 2.02\% & 24.722 & 35.867 & 60.588 & 1258ms & 1\% \\
  & \textbf{Ours, real} & \textbf{16.714} & \textbf{24.342} & \textbf{41.056} & \textbf{2116ms} & \textbf{0\%} & \textbf{23.814} & \textbf{29.763} & \textbf{53.577} & \textbf{2608ms} & \textbf{0\%} \\ \cline{2-12}
 & AM\cite{kool2018attention}, \revii{synthetic} & 17.687 & 27.778 & 45.465 & 1284ms & 2.02\% & 25.713 & 35.724 & 61.437 & 1264ms & 0\% \\
 & POMO\cite{kwon2020pomo}, \revii{synthetic} & 14.563 & 23.673 & 38.235 & 876ms & 86.87\% & 20.954 & 28.061 & 49.015 & 904ms & 97\% \\
 & \textbf{Ours, \revii{synthetic}} & \textbf{17.200} & \textbf{24.645} & \textbf{41.845} & \textbf{2324ms} & \textbf{0\%} & \textbf{25.085} & \textbf{31.619} & \textbf{56.704} & \textbf{2564ms} & \textbf{0\%} \\ \hline\hline
 \multirow{8}{*}{50} & HIGHS(21000s)+CPLEX(60s) & 16.398 & 30.939 & 47.338 & 5.85h & 0\% & 23.097 & 34.792 & 57.890 & 5.85h & 0\% \\
 & HIGHS(21000s)+CPLEX(300s) & 13.501 & 23.251 & 36.752 & 5.92h & 0\% & 19.523 & 27.313 & 46.836 & 5.92h & 0\% \\ \cline{2-12}
  & \revii{NSGA-II\cite{deb2002fast}} & 18.536 & 24.057 & 42.593  & 3496.05s  & 20\% & 28.121 & 33.892 & 62.013 & 3028.11s & 74\% \\
 & \revii{MOEA/D\cite{zhang2007moea}} & - & - & - & - & 100\% & - & - & - & - & 100\%  \\ \cline{2-12}
 & AM\cite{kool2018attention}, real & 13.091 & 27.929 & 41.020 & 2198ms & 0\% & 20.927 & 31.724 & 52.651 & 2285ms & 0\% \\
 & POMO\cite{kwon2020pomo}, real & 11.088 & 24.753 & 35.841 & 2077ms & 30\% & 19.237 & 29.359 & 48.596 & 2029ms & 28\% \\
 & AMDKD-AM\cite{bi2022learning} & 13.320 & 26.893 & 40.214 & 2345ms & 0\% & 20.359 & 32.234 & 52.593 & 2301ms & 0\% \\
  & \textbf{Ours, real} & \textbf{14.676} & \textbf{24.071} & \textbf{38.746} & \textbf{3996ms} & \textbf{0\%} & \textbf{21.738} & \textbf{30.130} & \textbf{51.868} & \textbf{2794ms} & \textbf{0\%} \\ \cline{2-12}
 & AM\cite{kool2018attention}, \revii{synthetic} & 14.158 & 27.928 & 42.086 & 2595ms & 0\% & 22.867 & 35.149 & 58.017 & 2436ms & 0\% \\
 & POMO\cite{kwon2020pomo}, \revii{synthetic} & 13.239 & 27.485 & 40.724 & 2063ms & 4\% & 20.093 & 31.633 & 51.726 & 2142ms & 19\% \\
 & \textbf{Ours, \revii{synthetic}} & \textbf{16.231} & \textbf{25.186} & \textbf{41.416} & \textbf{2701ms} & \textbf{0\%} & \textbf{23.937} & \textbf{32.716} & \textbf{56.652} & \textbf{2924ms} & \textbf{0\%} \\ \hline
\end{tabularx}
\end{table*}

\subsubsection{Evaluation of the Solution Quality}
\begin{table}[t]
    \centering
    \caption{Feasibility Rate and \revii{Execution Time} per Problem by Iteration}
    \label{tab:iter_process}
    \begin{tabular}{l|l|c|c|c|c}
    \hline
    \multicolumn{1}{c|}{} & \multirow{2}{*}{$N_U$} & \multicolumn{2}{c|}{Gunma} & \multicolumn{2}{c}{Osaka} \\
    \cline{3-6} 
    \multicolumn{1}{c|}{} & & \makecell{Feasibility \\ Rate \%} & \makecell{\revii{Execution} \\ \revii{Time[ms]}} & \makecell{Feasibility \\ Rate \%} & \makecell{\revii{Execution} \\ \revii{Time[ms]}} \\
    \hline
    \multirow{3}{*}{Iteration 1} & 20 & 77.6 & 984 & 73.8 & 1003 \\
     & 30 & 82.2 & 1564 & 79.4 & 1415 \\
     & 50 & 96 & 2585 & 96.6 & 2655 \\
    \hline
    \multirow{3}{*}{Iteration 2} & 20 & 90.4 & 1132 & 86.4 & 1138 \\
     & 30 & 93.7 & 1754 & 91.8 & 1612 \\
     & 50 & 99 & 2709 & 99.4 & 2754 \\
    \hline
    \end{tabular}
\end{table}
\rev{
We show the results of our evaluation in Table \ref{tab:results}. For our method, we set the embedding vector dimension to 128 in this experiment. \revii{Although a higher dimension (e.g., 256) tends to achieve better performance as discussed in Section \ref{sec:configparam}, we adopt 128 here because the computational cost increases as the number of dimensions gets larger, and it is extremely costly to evaluate using that settings across methods.}

The results obtained by the optimization solver show that the \revii{operating time} decreases as the execution time increases. While this difference is slight for $N_U=20$, a gap of over 10\% emerges for larger $N_U$ in both regions when comparing a 1-minute runtime to a longer runtime. This implies using data from longer runtimes for training could further improve the performance of our proposed method, indicating that our approach described below has strong potential for enhancement compared to RL methods, which do not incorporate expert knowledge.

As for constraint violation related to the vehicle number, while other methods frequently produce violations, our method satisfies all constraints except for one case ($N_U=20$, Osaka Facility). This demonstrates that our approach is more robust to constraint violations.
Furthermore, compared with AM and AMDKD-AM, which exhibit few constraint violations, our method improves the \revii{operating time} across all $N_U$ categories and regions. Especially, ours achieves gains of up to 8\% for $N_U=20$ and $N_U=30$ at the Gunma Facility. Although the \revii{execution time} is somewhat longer, \revii{mainly due to multiple iterations of inference and the heuristic procedure}, the difference remains within approximately 2 seconds per instance even in the worst case, relative to existing ML-based methods. We also confirmed that nearly 90\% of all instances are solved by the second iteration for every $N_U$ and regions; thus, the increased runtime is largely attributable to a small number of hard instances that require more iterations to obtain a feasible solution. As shown in Table \ref{tab:iter_process}, the \revii{execution time} up to the second iteration is almost indistinguishable from that of existing methods. These results confirm that our approach achieves a favorable balance between accuracy, computational efficiency, and robustness to constraint violations.

\revii{
Regarding the GA-based methods, NSGA-II achieved better operating times than the proposed method for some numbers of users. However, the violation rate increased for larger numbers of users, and execution times became excessively long. Furthermore, MOEA/D was unable to produce feasible solutions for all users and regions. This implies that GA-based methods are not practical for real-world applications.}

On the other hand, there is still room for improvement in terms of \revii{execution time} and solutions' quality.
Our method exhibits the longest execution time among all compared ML-based methods, mainly due to the overhead introduced by the post-processing refinement and the additional iterations required when a single refinement step fails to satisfy the vehicle number constraint. Although this post-processing ensures feasibility, it increases computation time up to five times longer than POMO, the fastest baseline.
As for the \revii{operating time}, the gap between our method and existing methods becomes small for $N_U=50$ compared to smaller $N_U$, indicating that our current \revii{learning model's} performance varies with the number of users and that it has not yet achieved full generalization. Future work should focus on developing more robust methods to user size variations.

}

\subsubsection{Quality Comparison Across Different Training Data Types}
As shown in Table \ref{tab:results}, training with real-world data generally yields higher-quality solutions than using \revii{synthetic data}.
However, there is a trade-off between privacy protection and usability: it is risky to use real data on the real-world service because of the possibility of personal data leakage, while we get lower-quality solutions when training with \revii{synthetic data} than real data. \revii{In the future, we will aim to develop a privacy-aware learning framework, for example by incorporating privacy protection mechanisms into the data or the learning model}.

\rev{
\subsubsection{Configuration of the parameters} \label{sec:configparam}
\begin{table}[t]
    \centering
    \caption{\revii{Operating times} across different embedding dimensions}
    \label{tab:difembed}
    \begin{tabular}{c|c|cc|cc}
    \hline
    \multirow{2}{*}{$N_U$} & \multirow{2}{*}{$d_h$} & \multicolumn{2}{c|}{Gunma} & \multicolumn{2}{c}{Osaka} \\
    \cline{3-6}
     & & \makecell{OT[min] \\(real)} & \makecell{OT[min] \\(\revii{synthetic})} & \makecell{OT[min] \\(real)} & \makecell{OT[min] \\(\revii{synthetic})} \\
    \hline
    \multirow{3}{*}{20} & 64 & 38.859 & 40.925 & 52.405 & 55.890 \\
    
     & 128 & \textbf{38.332} & 40.794 & \textbf{52.252} & 55.759 \\
    
     & 256 & 38.655 & \textbf{40.758} & 52.374 & \textbf{54.848} \\
    \hline
    \multirow{3}{*}{30} & 64 & 41.009 & 41.966 & 54.223 & 56.973 \\
    
     & 128 & 41.056 & 41.845 & 53.577 & 56.704 \\
    
     & 256 & \textbf{40.835} & \textbf{41.511} & \textbf{53.411} & \textbf{56.553} \\
    \hline
    \multirow{3}{*}{50} & 64 & 39.142 & 41.696 & 52.281 & 56.780 \\
    
     & 128 & 38.746 & 41.416 & 51.868 & 56.652 \\
    
     & 256 & \textbf{38.491} & \textbf{40.933} & \textbf{50.903} & \textbf{56.479} \\
    \hline
    \end{tabular}
\end{table}
We present in Table \ref{tab:difembed} the results obtained by varying the embedding dimension. As shown in the table, a larger embedding dimension generally leads to a smaller \revii{operating time}.
When the number of users increases, the dimensionality of the input vectors also becomes larger, and therefore a higher-dimensional feature space is required to preserve all relevant information.
In addition, a larger number of users implies a greater number of candidate locations to visit, which requires the embedding vectors to represent the differences among nodes more distinctly.
Higher-dimensional embeddings can better preserve such differences between vectors, thereby contributing to more accurate predictions of the ground truth.
Taken together, these observations suggest that choosing a larger embedding dimension is preferable, particularly when dealing with instances that involve a large number of users.
}


%% file: 06_conclusion.tex
\section{Conclusion}
In this paper, we investigated the optimization of Nursing Care Taxi Dispatch (NCTD) using machine learning to provide mobility support through AI-based decision-making that leverages various personal data.
We defined the necessary constraints for the car dispatch of the nursing homes, focusing on older users’ attributes, and then formulated NCTD as a mixed integer linear programming problem based on the Vehicle Routing Problem.
Furthermore, based on an existing method using Attention mechanism, we proposed a new method to solve NCTD in a short time.
\rev{Our method reduces \revii{operating times} by up to 8\% compared with existing approaches, while also avoiding violations of vehicle number constraints.
Moreover, in many cases, it succeeds in producing solutions within execution times comparable to those of the existing methods.}
Currently, we have a critical issue with the proposed method: there is a trade-off between the quality of training data and the time required to generate them. We need a significant amount of time to generate labels of the true trajectories to create enough amount of learning data. Therefore, we plan to introduce new learning methods, especially unsupervised learning techniques such as contrastive learning or weakly supervised learning.
Additionally, we aim to develop a privacy-preserving AI model that can securely handle sensitive personal data, thereby enabling practical deployment in real-world care facilities.